\documentclass[lettersize,journal]{IEEEtran}
\usepackage{algorithm}
\usepackage{algpseudocode}
\usepackage{epsfig}
\usepackage{graphicx}
\usepackage{amsfonts}
\usepackage{comment}
\usepackage{caption}
\usepackage{subcaption}
\usepackage{color}
\usepackage{amssymb}
\usepackage{amsmath}
\usepackage{adjustbox}
\usepackage{amsthm,mathtools,mathrsfs}
\usepackage{multirow}
\usepackage{etoolbox}
\usepackage{balance}
\usepackage{booktabs}
\usepackage{enumitem}
\usepackage{pifont,booktabs}
\usepackage[table]{xcolor}
\usepackage[colorlinks = true,
    linkcolor = red,
    urlcolor  = black,
    citecolor = blue, 
    anchorcolor = magenta]{hyperref}

\definecolor{altMixTen}{HTML}{3978B6}
\definecolor{altMixTwenty}{HTML}{E5A04A}
\definecolor{altMixHighlight}{HTML}{EAF4EF}

\usepackage{tikz}

\usepackage{caption}
\usepackage{subcaption}
\usepackage{boldline}
\usepackage{makecell}
\setcellgapes{1pt}

\usepackage{lineno}
\usepackage{adjustbox}
\usepackage{dsfont}

\usepackage{pgfplots}
\pgfplotsset{compat=newest}
\usepgfplotslibrary{groupplots}
\definecolor{cWarm}{HTML}{1F77B4}
\definecolor{cTen}{HTML}{FF7F0E}
\definecolor{cTwenty}{HTML}{2CA02C}
\pgfplotsset{sensrow/.style={
    scale only axis, width=0.195\textwidth, height=0.15\textwidth,
    xmin=0.035, xmax=0.315, xtick={0.05,0.1,0.15,0.2,0.25,0.3}, xticklabels={0.05,0.1,0.15,0.2,0.25,0.3},
    xticklabel style={font=\scriptsize}, yticklabel style={font=\scriptsize, /pgf/number format/fixed, /pgf/number format/precision=3},
    scaled y ticks=false, xlabel={Dropout rate $p$},
    label style={font=\scriptsize}, title style={font=\footnotesize, yshift=-4pt},
    grid=major, grid style={black!12},
    legend style={font=\scriptsize, draw=none, legend columns=-1, /tikz/every even column/.append style={column sep=8pt}}}}

\AfterEndEnvironment{lemma}{\noindent\ignorespaces}

\newtheorem{definition}{Definition}[section]
\AfterEndEnvironment{definition}{\noindent\ignorespaces}

\AfterEndEnvironment{example}{\noindent\ignorespaces}

\ifCLASSINFOpdf
\else
\fi
\begin{document}

\title{SHARP-QoS: \underline{S}parsely-gated \underline{H}ierarchical \underline{A}daptive \underline{R}outing for Joint \underline{P}rediction of QoS}

\author{Suraj Kumar,~\IEEEmembership{Graduate Student Member,~IEEE},
    Arvind Kumar,
    Soumi Chattopadhyay,~\IEEEmembership{Senior Member,~IEEE}
\thanks{
Suraj Kumar, Arvind Kumar, and Soumi Chattopadhyay are with the Dept. of CSE, Indian Institute of Technology Indore, India 453552. 
\\
Corresponding author: Soumi Chattopadhyay (email: soumi@iiti.ac.in)
}
}


\markboth{IEEE Transactions on Services Computing}{SHARP-QoS}


\maketitle


\begin{abstract}
Dependable service-oriented computing relies on multiple Quality-of-Service (QoS) parameters to assess service performance. However, real-world QoS data are often sparse, noisy, and shaped by hierarchical dependencies arising from QoS interactions, geographical constraints, and network-level factors, making accurate prediction challenging. Existing deep learning-based methods often predict each QoS parameter separately, requiring multiple similar models, which increases computational overhead and limits their suitability for applications requiring joint decisions across multiple QoS parameters. Although recent joint QoS prediction studies employ shared architectures, differences in numerical ranges across QoS parameters can lead to task imbalance and negative transfer, while inadequate representation learning and suboptimal joint optimization can degrade prediction accuracy. We present SHARP-QoS, a joint QoS prediction framework with three components. First, hyperbolic graph and hypergraph convolutions in the Poincar\'{e} ball extract hierarchical features from QoS invocation and contextual graphs. Second, sparsely gated sub-network routing selectively shares contextual and cross-QoS features, and a gated fusion module combines them with the structural features. Third, an EMA-based loss-balancing strategy smooths short-term fluctuations in task losses to provide less fluctuating task weights. Experiments on three multi-QoS datasets covering two, three, and four QoS parameters show that SHARP-QoS achieves lower prediction errors than existing joint QoS prediction methods. Because its contextual branch relies only on public region and AS attributes, it supports prediction for new users and services. Ablation, hyperparameter, outlier, and scalability analyses further assess its performance, highlighting the trade-off between prediction accuracy and computational cost.
\end{abstract}


\section{Introduction}\label{sec:intro}

Service-oriented architectures (SoA) build web and cloud applications from independent, reusable, and loosely coupled services \cite{zibinSurvey2022}. As many providers now offer functionally similar services, choosing among them depends on QoS parameters such as response time, throughput, and reliability, which describe the non-functional behavior of services and are widely used for service selection, composition, and recommendation~\cite{ghafouriSurvey2022}. These parameters are often considered together: a service chosen for a latency-sensitive workflow must meet its response-time, throughput, and reliability targets simultaneously. Since monitoring every user-service pair is costly, QoS prediction is used to estimate unobserved values \cite{cmf_www21, ncrl_2023_tsc}.

Traditional QoS prediction methods rely on collaborative filtering (CF) over user and service similarities \cite{upcc_1998_uai, ipcc_www_2001, wsrec_2011_tsc}, but remain sensitive to data sparsity, cold-start cases, and outliers. Matrix factorization (MF) \cite{cmf_www21,cnmf,pmf_nips_2007} and factorization machines (FM) \cite{EFMPred} partly address these issues by learning latent interactions. Deep models, including multilayer perceptrons (MLPs) \cite{ncrl_2023_tsc,MM-DNN} and graph convolutional networks (GCNs) \cite{QoSGNN_TSC_2024, llmQoS_2025_sse}, capture nonlinear dependencies and graph structure, while hybrid methods combine these approaches \cite{ncrl_2023_tsc, offdq, arrqp_tsc_2025}. Most of these methods, however, predict one QoS parameter at a time. Predicting multiple parameters therefore requires one separately trained model per parameter, which prevents sharing information about common users and services.

Multi-task learning (MTL), formalized by \cite{caruana1997multitask}, learns multiple tasks jointly by sharing information across them. General MTL architectures such as Cross-Stitch~\cite{misra2016_cross-stich}, MMoE~\cite{ma2018_MMoE}, and PLE~\cite{tang2020_PLE} share features through learned combinations or expert gates. Recent studies have applied MTL to joint QoS prediction. DNM~\cite{DNM_2021_TSC} models cross-context interactions through additive and multiplicative operations, while JQSP~\cite{JQSP_2023_TNSM} combines graph attention with an outlier-aware loss. MGEN~\cite{MGEN_2023_JP} uses gated experts to learn context- and QoS-specific representations, and HTG~\cite{HTT_2024_ETT} addresses cold-start cases through graph-based MTL but relies heavily on topology information. PMT~\cite{PMT_2023_TNSM} and WAMTL~\cite{WAMTL_2024_ICWS} balance tasks using uncertainty-based weighting and dynamic weight averaging, respectively. PLHEN~\cite{PLHEN_ESA_2026} uses hierarchical expert layers to adaptively combine shared and task-specific representations.

Despite these advances, three limitations remain. First, existing methods rarely capture hierarchical dependencies in QoS and contextual data while accounting for contextual correlations across QoS parameters. Second, feature sharing across QoS parameters is either limited or implemented through expert-based architectures that increase complexity~\cite{MGEN_2023_JP,PLHEN_ESA_2026}. Third, QoS parameters can differ in numerical range (Table~\ref{tab:datasets}), allowing larger-scale targets to dominate an unweighted joint loss and potentially impair joint learning. Existing balancing strategies based on uncertainty or dynamic loss changes~\cite{PMT_2023_TNSM,WAMTL_2024_ICWS} may also produce fluctuating task weights when losses vary over short periods.

To address these limitations, we propose SHARP-QoS, a joint QoS prediction framework with three complementary components:
\textit{(i) Hierarchical Feature Extraction Block (HFEB):} applies hyperbolic graph convolution to region and AS graphs and hyperbolic hypergraph convolution to QoS invocation graphs. The structural characteristics of these graphs are examined in Appendices~E--G.
\textit{(ii) Two-level feature sharing:} a contextual router based on Sub-Network Routing (SNR)~\cite{SNR} assigns shared contextual features to each QoS task. We further introduce Cross-SNR, a routing mechanism that enables information exchange between the structural features of different QoS tasks. A gated fusion module combines these shared representations with each task's structural features.
\textit{(iii) EMA-based loss balancing:} assigns task weights using the inverse of each task's exponentially averaged loss, smoothing short-term fluctuations in the loss estimates used for weighting.

The contextual branch relies only on public region and AS attributes, enabling cold-start prediction for new users and services with known contextual attributes, without requiring user- or service-specific identifiers.

The main \textbf{contributions} are summarized as follows:
\begin{itemize}[left=0pt]
    \item {We propose SHARP-QoS, a joint QoS prediction framework combining hyperbolic graph and hypergraph convolution with two-level sub-network routing and gated fusion, using coarse-grained contextual attributes and QoS-aware structural representations.}

    \item {We adopt EMA-based inverse-loss weighting to mitigate scale imbalance and smooth task-weight fluctuations, and compare it with gradient-based and loss-scaling strategies.}

    \item {We evaluate SHARP-QoS on WSDREAM-2T~\cite{wsdream}, Small-3T~\cite{PMT_2023_TNSM}, and gRPC-4T~\cite{arrqp_tsc_2025}, covering two, three, and four QoS parameters. Our evaluation includes ablation, cold-start, outlier sensitivity, hyperparameter sensitivity, scalability, and confidence interval analyses. We also examine the graph hierarchy assumption using Gromov $\delta$-hyperbolicity, Ollivier-Ricci curvature, and embedding distortion.}
\end{itemize}
The remainder of the paper is organized as follows: hyperbolic preliminaries (\({\S}~\)\ref{sec:preliminaries}), problem formulation (\({\S}~\)\ref{sec:problem_formulation}), the SHARP-QoS framework (\({\S}~\)\ref{sec:method}), experimental results (\({\S}~\)\ref{sec:experiment_section}), related work (\({\S}~\)\ref{sec:related}), and conclusions (\({\S}~\)\ref{sec:conclusion}).
\section{Preliminaries}\label{sec:preliminaries}

QoS invocation graphs exhibit tree-like structures, while region and AS groupings are hub-dominated (Appendices~E--F). Euclidean space struggles to represent such structures because trees grow exponentially with depth, whereas Euclidean volume grows polynomially. In contrast, hyperbolic space exhibits exponential volume growth, enabling low-distortion embeddings of hierarchical structures~\cite{hyperbolic_nn_nips_2018,hyperbolic_gcn_nips_2019}. We adopt the Poincar\'e ball for its conformal geometry and closed-form operations at the origin, and summarize the definitions used by HFEB (\S.~\ref{sec:hfeb}) below.

\subsection{Poincar\'e Ball Model}
The $d$-dimensional Poincar\'e ball with curvature $-c$ is
\(
    \mathbb{B}^{d,c}
    =\{\mathbf{x}\in\mathbb{R}^{d}\mid c\|\mathbf{x}\|^{2}<1\},
\)
where $c>0$ is the curvature magnitude.
Its conformal metric is
\(
g_{\mathbf{x}}=\lambda_{\mathbf{x}}^{2}g_{\mathrm{E}},
    \qquad
    \lambda_{\mathbf{x}}=\frac{2}{1-c\|\mathbf{x}\|^{2}},
\)
where $g_{\mathrm{E}}$ is the Euclidean metric tensor.
Distances increase sharply near the boundary, providing room
to represent hierarchical relationships.
In SHARP-QoS, we use a fixed curvature ${-}1$. This provides a consistent
hyperbolic geometry across the graph and hypergraph convolution layers.

\subsection{Hyperbolic Operations at the Origin}

We map between the Poincar\'e ball and its tangent space at the origin $\mathbf{0}$, where feature aggregation can use standard matrix operations~\cite{hyperbolic_gcn_nips_2019}. Each hypergraph convolution layer (HyConv) (Eq.~\ref{eq:hgconv}) uses the following operations.

\begin{itemize}[left=0pt]
    \item \textbf{Exponential and logarithmic maps.} These maps transform representations between the ball and its tangent space:
    \begin{equation}
        \scriptsize
        \exp_{\mathbf{0}}^{c}(\mathbf{v})=
        \tanh(\sqrt{c}\|\mathbf{v}\|)
        \frac{\mathbf{v}}{\sqrt{c}\|\mathbf{v}\|},
        ~~
        \log_{\mathbf{0}}^{c}(\mathbf{x})=
        \frac{\operatorname{artanh}(\sqrt{c}\|\mathbf{x}\|)}{\sqrt{c}}
        \frac{\mathbf{x}}{\|\mathbf{x}\|}.
    \end{equation}
    HyConv aggregates neighbor features in the tangent space and maps the result back to the ball using $\exp_{\mathbf{0}}^{c}$.

    \item \textbf{M\"obius addition} ($\oplus_c$). This operation provides a hyperbolic counterpart to vector addition and supports bias translation:
    \begin{equation}
        \scriptsize
        \mathbf{x}\oplus_c\mathbf{y}=
        \frac{
        (1+2c\langle\mathbf{x},\mathbf{y}\rangle+c\|\mathbf{y}\|^2)\mathbf{x}
        +(1-c\|\mathbf{x}\|^2)\mathbf{y}
        }{
        1+2c\langle\mathbf{x},\mathbf{y}\rangle
        +c^2\|\mathbf{x}\|^2\|\mathbf{y}\|^2
        }.
    \end{equation}

    \item \textbf{M\"obius matrix--vector multiplication} ($\otimes_c$). This operation applies a learnable linear transformation in the tangent space and maps the result back to the ball:
    \begin{equation}
        \scriptsize
        \mathbf{W}\otimes_c\mathbf{x}
        =\exp_{\mathbf{0}}^{c}
        \big(\mathbf{W}\log_{\mathbf{0}}^{c}(\mathbf{x})\big).
    \end{equation}

    \item \textbf{Wrapped activation} ($\sigma^{\otimes_c}$). This operation applies a nonlinearity in the tangent space and maps the output back to the ball:
    \begin{equation}
        \scriptsize
        \sigma^{\otimes_c}(\mathbf{x})
        =\exp_{\mathbf{0}}^{c}
        \big(\sigma(\log_{\mathbf{0}}^{c}(\mathbf{x}))\big).
    \end{equation}
\end{itemize}

These operations allow SHARP-QoS to perform feature transformations and aggregation while respecting the Poincar\'e ball geometry. The learnable parameters can be optimized using a standard Euclidean optimizer, without requiring a Riemannian optimizer.
\section{Problem Formulation}\label{sec:problem_formulation}
Let $\mathcal{U}=\{u_1,\dots,u_n\}$ and $\mathcal{S}=\{s_1,\dots,s_m\}$ denote the sets of $n$ users and $m$ services. Each user and service is described only by two coarse, public attributes: its geographical region (RG) and autonomous system (AS).

For $\mathrm{P}$ QoS parameters, let $\mathbf{Q}^{p}\in\mathbb{R}^{n\times m}$, $p\in\{1,\dots,\mathrm{P}\}$, be a partially observed user--service matrix:
\begin{equation}\label{eq:qos_def_rewrite}
    \mathbf{Q}^{p}_{ij} =
    \begin{cases}
    q^{p}_{ij} > 0, & \text{if user $u_i$ has invoked service $s_j$},\\
    0, & \text{otherwise},
    \end{cases}
\end{equation}
and let $\Omega^{p}=\{(i,j)\mid \mathbf{Q}^{p}_{ij}>0\}$ be the set of observed entries. Given $\{\mathbf{Q}^{p}\}_{p=1}^{\mathrm{P}}$ and the contextual attributes, the goal is to learn a single model that predicts $\hat{\mathbf{Q}}^{p}_{ij}$ for all $\mathrm{P}$ parameters and all unobserved pairs $(i,j)\notin\Omega^{p}$. This includes cold-start users and services, which have no observed entries and are described only by their contextual attributes.

\begin{figure*}[!t]
    \centering
    \includegraphics[width=0.85\linewidth]{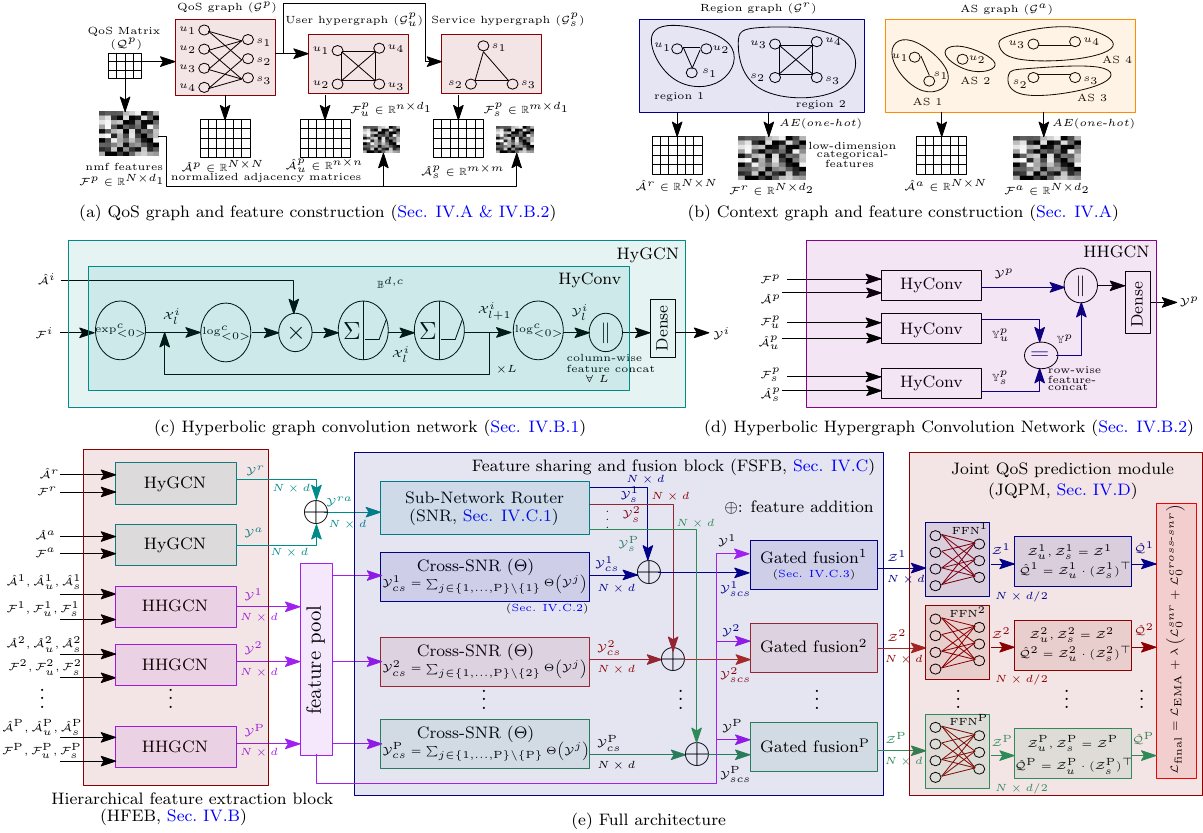}
    \caption{SHARP-QoS: Overall framework.}
    \label{fig:arch}
\end{figure*}

\begin{table}[!b]
    \centering
    \scriptsize
    \caption{Key notations.}
    \label{tab:notation}
    \begin{tabular}{l|l}
        \hline
        \textbf{Symbol} & \textbf{Description} \\ \hline
        $\mathcal{U}, \mathcal{S}$ & Sets of users and services \\
        $n, m, N$ & Numbers of users, services, and nodes ($N=n+m$) \\
        $\mathrm{P}$ & Number of QoS parameters \\
        $\mathbf{Q}^p,\ \hat{\mathbf{Q}}^p$ & Observed and predicted QoS matrices for parameter $p$ \\
        $\mathcal{G}^p$ & QoS invocation graph for parameter $p$ \\
        $\mathcal{G}^p_u, \mathcal{G}^p_s$ & User and service hypergraphs for parameter $p$ \\
        $\mathcal{G}^r, \mathcal{G}^a$ & Region and AS context graphs \\
        $\mathbb{B}^{d,c}$ & $d$-dimensional Poincar\'e ball with curvature $-c$ \\
        $d,\ L$ & Hidden dimension and number of HyConv layers \\
        $K_1, K_2$ & Numbers of SNR and Cross-SNR blocks \\
        \hline
    \end{tabular}
\end{table}

Table~\ref{tab:notation} summarizes the key notation.
 \section{Methodology}\label{sec:method}
Fig.~\ref{fig:arch} illustrates SHARP-QoS, comprising four stages: (i) \textbf{Preprocessing}, which constructs QoS and context graphs and initializes node features via matrix factorization and one-hot encoding; (ii) \textbf{Hierarchical Feature Extraction Block (HFEB)}, which learns QoS- and context-specific representations through hyperbolic graph and hypergraph convolution; (iii) \textbf{Feature Sharing and Fusion Block (FSFB)}, which adaptively shares features across QoS parameters via subnetwork routing and gating-based fusion; and (iv) \textbf{Joint QoS Prediction Module (JQPM)}, which jointly predicts all QoS parameters using task-specific feed-forward networks (FFNs) and a matrix product, with EMA-based loss balancing.

Hyperbolic computations follow the Poincaré ball formulation in \S~\ref{sec:preliminaries} and are confined to HFEB; FSFB and JQPM operate in Euclidean space.

\subsection{Preprocessing}\label{sec:preprocessing}
We first build the propagation graphs and initialize the node features from the QoS matrices and contextual attributes.

\subsubsection{QoS- and Context-Graph Construction}
Prior graph-based QoS predictors use QoS and context graphs with noisy or uninformative connections~\cite{QoSGNN_TSC_2024}. We instead build graphs whose edges come only from observed invocations and shared contextual attributes, defined as follows.

\begin{definition}[QoS Invocation Graph (\(\mathcal{G}^p\))]\label{def:invocation_graph}
Given a QoS matrix \(\mathbf{Q}^p\), \(p\in \{1,\dots, \mathrm{P}\}\), the invocation graph \(\mathcal{G}^p=(\mathcal{V},\mathcal{E}^p)\), with \(\mathcal{V}=\mathcal{U}\cup\mathcal{S}\), is a bipartite graph in which an edge \((u_i,s_j)\in\mathcal{E}^p\) exists if and only if \((i,j)\in\Omega^p_{\mathrm{tr}}\), i.e., user \(u_i\) has invoked service \(s_j\) in the training data.
\end{definition}

\begin{definition}[Context Graphs (\(\mathcal{G}^r,\mathcal{G}^a\))]\label{def:context_graph}
For each context attribute \(\nu\in\{r,a\}\), the context graph \(\mathcal{G}^\nu=(\mathcal{V},\mathcal{E}^\nu)\) connects every pair of entities that share the same region (\(r\)) or AS (\(a\)).
\end{definition}

The graph \(\mathcal{G}^p\) captures collaborative QoS usage patterns; when all QoS parameters are recorded for the same invocations, the \(\mathcal{G}^p\) coincide across \(p\). The graphs \(\mathcal{G}^r\) and \(\mathcal{G}^a\) capture coarse geographic proximity and finer-grained network proximity, respectively. Although many AS-level edges also appear in \(\mathcal{G}^r\), the two graphs offer complementary
views, and using both improves prediction (Table~\ref{tab:context_gconv}).

We represent each graph by its adjacency matrix \(\mathbf{A}^\nu\), \(\nu\in\{r,a,1,\dots,\mathrm{P}\}\), giving \(\mathrm{P}{+}2\) matrices of size \(N{{\times} N}\), where \(N=n+m\). Raw adjacency matrices in GCNs~\cite{gcn_iclr_2017} can cause numerical instability at high-degree nodes and ignore each node's own features. We therefore add self-loops and normalize symmetrically:
\begin{equation}
    \widehat{\mathbf{A}}^{\nu}
    =
    (\mathbf{D}^{\nu})^{-1/2}
    \mathbf{A}^{\nu}
    (\mathbf{D}^{\nu})^{-1/2}
    +\mathbf{I},
\end{equation}
where $\mathbf{D}^{\nu}$ is the degree matrix of $\mathbf{A}^{\nu}$. For isolated nodes, the corresponding inverse-degree entries are set to zero. The identity matrix introduces self-loops, ensuring that each node retains its own features during propagation.
\subsubsection{Initial Feature Extraction}\label{sec:initial_features}
We derive initial node features from the sparse QoS matrices and the contextual attributes (region, AS).

\textbf{(i) QoS-based features.} For each QoS parameter \(p\), we apply nonnegative matrix factorization (NMF)~\cite{nmf_nature_1999} to the training entries of \(\mathbf{Q}^p\), obtaining a user feature matrix \(\mathbf{F}^p_u \in \mathbb{R}^{n \times d_1}\) and a service feature matrix \(\mathbf{F}^p_s \in \mathbb{R}^{m \times d_1}\) such that
\(
    \mathbf{F}^p_u(\mathbf{F}^p_s)^\top \approx \mathbf{Q}^p, \text{where } \mathbf{F}^p_u,\mathbf{F}^p_s\ge 0.
\)
These low-rank factors provide each node with a dense initial representation despite the sparsity of \(\mathbf{Q}^p\).

\textbf{(ii) Context-based features.}
We one-hot encode the region and AS categories of each user and service. Since one-hot vectors grow with the number of categories, we compress them with four autoencoders (AE)~\cite{stacked_ae_jmlr_2010}, one each for user region, user AS, service region, and service AS, all with output dimension \(d_2\). This yields user features \(\mathbf{F}^{r}_{u},\mathbf{F}^{a}_{u}\in\mathbb{R}^{n\times d_2}\) and service features \(\mathbf{F}^{r}_{s},\mathbf{F}^{a}_{s}\in\mathbb{R}^{m\times d_2}\).

For each source \(i\in\{r,a,1,\dots,\mathrm{P}\}\), we stack the user and service features row-wise to form the node feature matrix \(\mathbf{F}^{i}=[\mathbf{F}^{i}_u;\mathbf{F}^{i}_s]\in\mathbb{R}^{N\times d}\), where \(d=d_1=d_2\).

\subsection{Hierarchical Feature Extraction Block (HFEB)}\label{sec:hfeb}
Previous approaches~\cite{QoSGNN_TSC_2024, JQSP_2023_TNSM, HTT_2024_ETT} use Euclidean graph convolution or attention (GCN/GAT) for neighborhood aggregation. However, Euclidean representations may distort tree-like structures in QoS graphs (Appendices~E-F). We therefore use the Poincar\'e ball \(\mathbb{B}^{d,c}\), with M\"obius transformations and nonlinearities (\S.~\ref{sec:preliminaries}). The HFEB employs separate hyperbolic encoders for contextual and QoS graphs, as described below.

\subsubsection{Hyperbolic Graph Convolution Network (HyGCN)}\label{subsubsec:hygcn}
Following~\cite{hyperbolic_nn_nips_2018, hyperbolic_gcn_nips_2019}, HyGCN extracts hierarchical features from the context graphs \(\mathcal{G}^r\) and \(\mathcal{G}^a\). Given an initial feature matrix \(\mathbf{F}^{i}\) and normalized adjacency \(\hat{\mathbf{A}}^{i}\), \(i\in\{r,a\}\), it first projects the features onto the manifold, \(\mathbf{X}^{i}_{0}=\exp_{\mathbf{0}}^{c}(\mathbf{F}^{i})\). Each hyperbolic convolution (HyConv) unit, shown in Fig.~\ref{fig:arch}(c), then performs two operations: message passing in the tangent space~\cite{gcn_iclr_2017}, followed by a M\"obius linear transformation with a wrapped ReLU activation:
\begin{equation}\label{eq:hgconv}
    \begin{aligned}
        \tilde{\mathbf{X}}^{i}_{l} &=
        \sigma_1^{\otimes_{c}}\!\left(
            \exp_{\mathbf{0}}^{c}\!\left(\hat{\mathbf{A}}^{i}\,\log_{\mathbf{0}}^{c}(\mathbf{X}^{i}_{l-1})\,\mathbf{W}^{i}_{l1}\right)
            \oplus_{c}\exp_{\mathbf{0}}^{c}(\mathbf{b}^{i}_{l1})
        \right),\\
        \mathbf{X}^{i}_{l} &=
        \sigma_1^{\otimes_{c}}\!\left(
            \exp_{\mathbf{0}}^{c}\!\left(\log_{\mathbf{0}}^{c}(\tilde{\mathbf{X}}^{i}_{l})\,\mathbf{W}^{i}_{l2}\right)
            \oplus_{c}\exp_{\mathbf{0}}^{c}(\mathbf{b}^{i}_{l2}) \right),
    \end{aligned}
\end{equation}
where \(l{\in}\{1,\dots,L\}\), \(\mathbf{W}^{i}_{l1},\mathbf{W}^{i}_{l2}\in\mathbb{R}^{d\times d}\) are learnable weights, \(\mathbf{b}^{i}_{l1},\mathbf{b}^{i}_{l2}\in\mathbb{R}^{d}\) are learnable biases.
The output of each layer is mapped back to Euclidean space as \(\mathbf{Y}^{i}_{l}=\log_{\mathbf{0}}^{c}(\mathbf{X}^{i}_{l})\in\mathbb{R}^{N\times d}\).

HyGCN stacks \(L\) HyConv layers, concatenates their outputs column-wise, and applies a nonlinear transformation with weights \(\mathbf{W}^i_1\in\mathbb{R}^{(L+1)d\times d}\) to obtain the hierarchical features:
\begin{equation}\label{eq:HyGCN_features}
    \mathbf{Y}^{i} = \sigma_1\!\left(\left(\Vert^{L}_{l=0}\, \mathbf{Y}^i_{l}\right)\mathbf{W}^i_1\right).
\end{equation}
Unlike Eq.~\eqref{eq:hgconv}, this readout operates in Euclidean space. Applying HyGCN to \(\mathcal{G}^{r}\) and \(\mathcal{G}^{a}\) yields contextual features \(\mathbf{Y}^{r},\mathbf{Y}^{a}\in\mathbb{R}^{N\times d}\), whose sum \(\mathbf{Y}^{ra}=\mathbf{Y}^{r}+\mathbf{Y}^{a}\) provides the geographic- and network-aware features used in \S.~\ref{subsec:fsfb}.

\subsubsection{Hyperbolic Hypergraph Convolution Network (HHGCN)}\label{subsubsec:hhgcn}
Applying HyGCN directly to the QoS graph \(\mathcal{G}^p\) yields weaker hierarchical features, because QoS interactions are sparse and noisy. We therefore complement the bipartite structure with user- and service-based hypergraphs. The user hypergraph \(\mathcal{G}^p_u\) is defined as follows.

{\begin{definition}[User Hypergraph (\(\mathcal{G}^p_u\))]\label{def:hypergraph}
Given a QoS invocation graph \(\mathcal{G}^p\), the user hypergraph \(\mathcal{G}^p_u=(\mathcal{U},\mathcal{E}^p_u)\) contains the \(n\) users as vertices and one hyperedge \(e_j\in\mathcal{E}^p_u\) for each service \(s_j\), \(j\in\{1,\dots,m\}\). Each hyperedge connects all users that invoked \(s_j\), i.e., the neighbors of \(s_j\) in \(\mathcal{G}^p\).
\end{definition}
}
{The hyperedges are represented by the incidence matrix \(\mathbf{H}^p\in\{0,1\}^{n\times m}\), where \(\mathbf{H}^p_{ij}=1\) if \(\mathbf{Q}^p_{ij}>0\) and \(0\) otherwise. The service hypergraph \(\mathcal{G}^p_s=(\mathcal{S},\mathcal{E}^p_s)\) is defined symmetrically, with one hyperedge per user connecting all services invoked by that user; its incidence matrix is \((\mathbf{H}^p)^\top\). Users that share a service, and services invoked by the same user, are thus connected through a common hyperedge, which exposes second-order co-invocation relations within a single convolution layer.} Following~\cite{HGNN_AAAI2019}, the normalized hypergraph adjacencies \(\hat{\mathbf{A}}^p_u\in\mathbb{R}^{n\times n}\) and \(\hat{\mathbf{A}}^p_s\in\mathbb{R}^{m\times m}\) are:
\begin{equation}\label{eq:hyper_adj}
    \begin{aligned}
        \hat{\mathbf{A}}^p_u &=
        (\mathbf{D}^p_u)^{-1/2}\mathbf{H}^p
        (\mathbf{D}^p_s)^{-1}(\mathbf{H}^p)^\top
        (\mathbf{D}^p_u)^{-1/2},\\
        \hat{\mathbf{A}}^p_s &=
        (\mathbf{D}^p_s)^{-1/2}(\mathbf{H}^p)^\top
        (\mathbf{D}^p_u)^{-1}\mathbf{H}^p
        (\mathbf{D}^p_s)^{-1/2},
    \end{aligned}
\end{equation}
where \(\mathbf{D}^p_u=\operatorname{diag}(\mathbf{H}^p\mathbf{1}_m)\) and \(\mathbf{D}^p_s=\operatorname{diag}((\mathbf{H}^p)^\top\mathbf{1}_n)\) are the user and service degree matrices, respectively. For entities without observed invocations, the corresponding inverse degree entries are set to zero (\(0^{-1}=0\)), so the normalization remains well-defined for cold-start entities.

HHGCN, shown in Fig.~\ref{fig:arch}(d), applies the HyConv layers of Eq.~\eqref{eq:hgconv} independently to \((\hat{\mathbf{A}}^p,\mathbf{F}^p)\), \((\hat{\mathbf{A}}^p_u,\mathbf{F}^p_u)\), and \((\hat{\mathbf{A}}^p_s,\mathbf{F}^p_s)\). Mapping the layer outputs back to Euclidean space gives \(\mathbf{Y}^p_l\in\mathbb{R}^{N\times d}\), \(\mathbf{Y}^p_{ul}\in\mathbb{R}^{n\times d}\), and \(\mathbf{Y}^p_{sl}\in\mathbb{R}^{m\times d}\), respectively. We stack \(\mathbf{Y}^p_{ul}\) and \(\mathbf{Y}^p_{sl}\) row-wise to align with the \(N\) nodes of \(\mathcal{G}^p\), concatenate the result column-wise with \(\mathbf{Y}^p_l\), and apply a nonlinear transformation with weights \(\mathbf{W}^p_2\in\mathbb{R}^{2(L+1)d\times d}\):
\begin{equation}\label{eq:hg_feat_fusion}
    \mathbf{Y}^p =
    \sigma_1\!\left(
    \left(\Vert^L_{l=0}\left(\mathbf{Y}^p_l \,\Vert\, \left[\mathbf{Y}^p_{ul} \,;\, \mathbf{Y}^p_{sl} \right]\right)\right) \mathbf{W}^p_2
    \right),
\end{equation}
where \(\sigma_1\) refers to ReLU. The output \(\mathbf{Y}^p \in \mathbb{R}^{N \times d}\) encodes the hierarchical QoS features of the \(p^{\text{th}}\) parameter.

\subsection{Feature Sharing and Fusion Block (FSFB)} \label{subsec:fsfb}

This block enables selective feature sharing across QoS parameters using representations learned by HyGCN and HHGCN. Hard parameter sharing can cause negative transfer, while expert-based approaches increase model complexity~\cite{WAMTL_2024_ICWS, PLHEN_ESA_2026}. To address these limitations, we employ sparse subnetwork routing at two levels: SNR~\cite{SNR} for task-specific contextual feature selection and the proposed Cross-SNR for cross-QoS structural feature sharing.

\subsubsection{Subnetwork Routing (SNR) Mechanism}\label{subsubsec:snr}

SNR~\cite{SNR} routes the contextual features \(\mathbf{Y}^{ra}\in\mathbb{R}^{N\times d}\) from HyGCN (\S~\ref{subsubsec:hygcn}) to each QoS parameter. The router processes \(\mathbf{Y}^{ra}\) through \(K_1\) parallel blocks \(\{\phi_k(\cdot)\}_{k=1}^{K_1}\), each comprising layer normalization (LN) followed by a ReLU-activated linear layer:
\begin{equation}\label{eq:dense_snr}
    \phi_k(\mathbf{Y}^{ra}) =
    \sigma_1\!\left(\mathrm{LN}(\mathbf{Y}^{ra})\,\mathbf{W}_k + \mathbf{b}_k\right)
    \in\mathbb{R}^{N\times d_o},
\end{equation}
where $\mathbf{W}_k\in\mathbb{R}^{d\times d_o}$ and $\mathbf{b}_k\in\mathbb{R}^{d_o}$ are the shared parameters of block $k$. We use $d=128$ and $d_o=64$. A QoS parameter-specific linear transformation $\mathbf{W}_k^{p}\in\mathbb{R}^{d_o\times d_o}$ then adapts each block's output:
\begin{equation}\label{eq:task-specific_transform}
    \tilde{\phi}_k^{p}(\mathbf{Y}^{ra}) =
    \phi_k(\mathbf{Y}^{ra})\,\mathbf{W}_k^{p}.
\end{equation}

Each QoS parameter \(p\) decides whether to use block \(k\) through a coding variable \(c_k^p\), drawn from a Bernoulli distribution \(\pi_k^p\). Because this variable is discrete, it does not allow gradient-based learning. We therefore use the hard concrete distribution~\cite{hardConcreteGate_iclr2018}, which parameterizes each \(c^{p}_k\) by a learnable logit \(\log \alpha_k^{p}\) and samples it using \(u \sim \mathcal{U}(0, 1)\):
\begin{equation} \label{eq:hard_concrete_gate}
    \begin{aligned}
        s &= \sigma_2\!\left(\left(\log u - \log(1-u) + \log \alpha_k^{p}\right)/\tau\right), \\
        c_k^{p} &= \mathrm{clip}(\bar{s}, 0, 1), \quad \bar{s} = s(\gamma_r - \gamma_l) + \gamma_l,
    \end{aligned}
\end{equation}
where \(\sigma_2\) is the sigmoid function, \(\gamma_l<0\) and \(\gamma_r>1\) define the stretching interval, and \(\tau\) controls the smoothness. The QoS parameter-specific contextual features are then obtained as a gated average of the blocks:
\begin{equation}\label{eq:snr_agg}
    \mathbf{Y}^{p}_{s} =  \left(\sum\nolimits_{k=1}^{K_1} c^{p}_{k}  \tilde{\phi}^{p}_{k} (\mathbf{Y}^{ra}) \right) /{\max \left(1,\ \sum\nolimits_{k=1}^{K_1} c^{p}_{k}\right)}  .
\end{equation}
The normalization keeps the feature scale stable when only a few blocks are active, and the floor of 1 keeps it defined when all gates are closed. At inference, the stochastic gates are replaced by deterministic ones, \(c_k^{p} = \mathbb{I}\!\left[\sigma_2(\log \alpha_k^{p}) > \delta\right]\) with threshold \(\delta\), so each QoS parameter uses a fixed, interpretable subset of blocks.

\subsubsection{Cross-SNR Routing Mechanism}\label{subsubsec:cross_snr}

QoS parameters share underlying network conditions but may exhibit different behavior. Unlike SNR~\cite{SNR}, which routes a shared input to individual tasks, Cross-SNR enables direct sharing of task-specific structural features. We propose Cross-SNR to selectively transfer QoS representations across tasks using the routing formulation in Eqs.~\eqref{eq:dense_snr}-\eqref{eq:snr_agg} with $K_2$ blocks. Given the HHGCN-derived feature pool $\{\mathbf{Y}^p\}_{p=1}^{\mathrm{P}}$ (Eq.~\eqref{eq:hg_feat_fusion}), the shared representation for target task $p$ aggregates routed features from the remaining $\mathrm{P}{-}1$ tasks, excluding its own:
\begin{equation}\label{eq:cross_snr}
    \mathbf{Y}^{p}_{cs} = \sum\nolimits_{j\in\{1,\dots,\mathrm{P}\}\setminus\{p\}} \Theta^{j\to p}\!\left(\mathbf{Y}^{j}\right),
\end{equation}
where \(\Theta^{j\to p}(\cdot)\) is a Cross-SNR router with its own \(K_2\) blocks, coding variables, and transforms, dedicated to routing the features of QoS parameter \(j\) to target \(p\). SHARP-QoS therefore uses \(\mathrm{P}(\mathrm{P}{-}1)\) Cross-SNR routers.

The combined shared features \(\mathbf{Y}^p_{scs} = \mathbf{Y}^p_{s} + \mathbf{Y}^p_{cs}\) sum the task-specific contextual and QoS features, and therefore carry network, geographic, and cross-QoS information.

\subsubsection{Gated Feature Fusion Module}\label{subsec:gated_feat_fusion}
This module adaptively combines 
the projected structural features ${\mathbf{Y}}^p\in
\mathbb{R}^{N\times d_o}$ and shared representations
$\mathbf{Y}_{scs}^p\in \mathbb{R}^{N\times d_o}$ are combined using a learnable gate $\mathbf{g}^p\in(0,1)^{N\times d_o}$, with $\mathbf{W}_g^p\in \mathbb{R}^{2d_o\times d_o}$:
\begin{align}
    \mathbf{g}^p &= \sigma_2\!\left([\mathbf{Y}^{p} \Vert \mathbf{Y}^{p}_{scs}]\,\mathbf{W}^p_{g}\right), \label{eq:gate_learning}\\
    \mathbf{Z}^{p} &= \mathbf{g}^p \odot \mathbf{Y}^{p} + (1-\mathbf{g}^p)\odot \mathbf{Y}^{p}_{scs}, \label{eq:gated_fusion}
\end{align}
where \(\odot\) denotes element-wise multiplication. The fused features \(\mathbf{Z}^p\) are passed to the prediction module.

\subsection{Joint QoS Prediction Module (JQPM)}\label{subsec:jqpm}

For each QoS parameter $p$, a task-specific two-layer feed-forward network maps the 64-dimensional fused representation to a 128-dimensional hidden space with ReLU activation, followed by a linear projection back to 64 dimensions. The resulting node embeddings are split into user and service representations, whose matrix product yields the QoS predictions:
\begin{equation}\label{eq:prediction}
    \bar{\mathbf{Z}}^{p} = \mathrm{FFN}^p(\mathbf{Z}^{p}) = \big[\bar{\mathbf{Z}}_u^p;\, \bar{\mathbf{Z}}_s^p\big], \qquad
    \hat{\mathbf{Q}}^{p} = \bar{\mathbf{Z}}_u^p\,(\bar{\mathbf{Z}}_s^p)^\top.
\end{equation}

\subsubsection{Objective Function and Loss Scale Balancing}\label{subsubsec:objective}
Network instability introduces outliers into QoS data, so we train with the mean absolute error (MAE), which is less sensitive to extreme values than the squared error:
\begin{equation}\label{eq:loss_fn}
    \mathcal{L}^p =
    \frac{1}{|\Omega^p_{\mathrm{tr}}|}\sum_{(i,j)\in \Omega^p_{\mathrm{tr}}} \big|\mathbf{Q}^{p}_{ij} - \hat{\mathbf{Q}}^{p}_{ij}\big|,
\end{equation}
where \(\Omega^p_{\mathrm{tr}}\subseteq\Omega^p\) is the set of observed training entries, so only observed QoS values contribute to the loss.

Because QoS parameters have very different numerical ranges (Table~\ref{tab:datasets}), the unweighted joint loss \(\sum_{p=1}^\mathrm{P}\mathcal{L}^p\) is dominated by the parameters with the largest errors, which can cause negative transfer across tasks. PMT~\cite{PMT_2023_TNSM} and WAMTL~\cite{WAMTL_2024_ICWS} rescale task losses with heteroscedastic uncertainty weighting (HUW) and dynamic weight averaging (DWA), respectively, but these schemes can be unstable in early training and adapt slowly to QoS dynamics. We instead use an EMA-based loss scaling strategy that balances task contributions while suppressing short-term fluctuations.

At the \(t^\text{th}\) training iteration, we compute the smoothed loss \(\tilde{\mathcal{L}}^p_{t}\) with a smoothing coefficient \(\beta \in [0,1)\) that controls the moving-average momentum, and obtain normalized task weights \(w^p_t\) by inverting the smoothed losses:
\begin{equation}\label{eq:ema_weight}
    \tilde{\mathcal{L}}^p_{t} = \beta \, \tilde{\mathcal{L}}^p_{t-1} + (1 - \beta) \, \mathcal{L}^p_t,~~
    w^p_t = \frac{(\tilde{\mathcal{L}}^p_t + \epsilon)^{-1}}{\sum_{q=1}^\mathrm{P} (\tilde{\mathcal{L}}^q_t +\epsilon)^{-1}},
\end{equation}
where \(\tilde{\mathcal{L}}^p_0 {=} 1\) and \(\epsilon\) is a small constant for numerical stability. Because each weight is inversely proportional to the task's typical loss, every weighted term \(w^p_t\mathcal{L}^p_t\) contributes on a comparable scale, regardless of the parameter's numerical range. The joint loss is \(\mathcal{L}_{\mathrm{EMA}} = \sum_{p=1}^\mathrm{P} w^p_t\, \mathcal{L}^p_t\).

To encourage sparse block selection in the routers, we add the expected \(L_0\) norm of the coding variables~\cite{hardConcreteGate_iclr2018}, \(\mathcal{R}_0=\sum_{p}\sum_{k}\sigma_2\big(\log\alpha_k^p-\tau\log(-\gamma_l/\gamma_r)\big)\), computed separately over the SNR (\(\mathcal{R}_0^{s}\)) and Cross-SNR (\(\mathcal{R}_0^{cs}\)) gates. The final objective is:
\begin{equation}\label{eq:final_loss}
    \mathcal{L}_{\mathrm{final}} = \mathcal{L}_{\mathrm{EMA}} + \lambda \left(\mathcal{R}_0^{s} + \mathcal{R}_0^{cs}\right),
\end{equation}
where the regularization coefficient \(\lambda\) controls the trade-off between prediction accuracy and routing sparsity. The computational complexity of SHARP-QoS is analyzed in Appendix~A of the supplementary file.
\section{Experimental Results}\label{sec:experiment_section}
SHARP-QoS is implemented in PyTorch 2.14.0 (CUDA 13.0) with Python 3.12.14. All experiments run on Ubuntu 26.04.1 LTS (Linux kernel 7.0.0) with an Intel Core Ultra 7 265 CPU, 16 GB of RAM, and an NVIDIA GeForce RTX 5060 Ti GPU with 16 GB of memory. 

\subsection{Experimental Setup}\label{sec:experiment_setup}

\textbf{Datasets.}
We use three real-world QoS datasets (statistics in Table~\ref{tab:datasets}):
(i) \emph{WSDREAM-2T}~\cite{wsdream}: response time (RT) and throughput (TP) for 339 users and 5,825 services;
(ii) \emph{Small-3T}~\cite{PMT_2023_TNSM}: RT, TP, and reliability (RE) for 112 users and 36 services; and
(iii) \emph{gRPC-4T}~\cite{arrqp_tsc_2025}: RE, cost (CT), latency (LT), and power (PW) for 57 users and 150 autonomous-vehicle gRPC services deployed on five academic servers.
All three provide the region and AS of each user and service.

{\textbf{Data splits.}
For each training density $\mathrm{TD}\in\{5,10,15,20\}\%$, we randomly sample TD\% of the observed records for training and use the rest for testing. 
We reserve $5\%$ of the training observations for validation and evaluate the model every 50 epochs. Validation observations are excluded from the training loss, graph construction, and NMF feature extraction. Model selection uses validation MAE, with early stopping based on validation performance.
For cold-start experiments (\S.~\ref{sec:coldstart}), we additionally remove all training records of a fraction of users and services, while keeping the test set unchanged. All methods are trained on the same data splits; each baseline uses the hyperparameters and validation protocol of its original paper, and every result is averaged over five random seeds.}

\textbf{Evaluation metrics.}
We report the mean absolute error (MAE) and root mean square error (RMSE)~\cite{cmf_www21,DNM_2021_TSC} over the test set $\Omega^p_{\mathrm{te}}$:
\begin{equation}\label{eq:metrics}
    \begin{aligned}
        \mathrm{MAE} &= \frac{1}{|\Omega^p_{\mathrm{te}}|}
        \sum_{(i,j)\in\Omega^p_{\mathrm{te}}}
        \big|\mathbf{Q}^p_{ij}-\hat{\mathbf{Q}}^p_{ij}\big|,\\
        \mathrm{RMSE} &= \sqrt{\frac{1}{|\Omega^p_{\mathrm{te}}|}
        \sum_{(i,j)\in\Omega^p_{\mathrm{te}}}
        \big(\mathbf{Q}^p_{ij}-\hat{\mathbf{Q}}^p_{ij}\big)^2}.
    \end{aligned}
\end{equation}
Lower values are better for both. We also report the relative improvement of SHARP-QoS over the strongest baseline,
\begin{equation}\label{eq:per_improvement}
    I = ({(e_{\mathrm{b}} - e_{\mathrm{ours}})}/{e_{\mathrm{b}}})\times 100\%,
\end{equation}
where $e_{\mathrm{ours}}$ and $e_{\mathrm{b}}$ are the MAE or RMSE of SHARP-QoS and of the strongest baseline, respectively. The hyperparameter configuration is given in Appendix~B of the supplementary file.

\begin{table}[!t]
    \centering
    \caption{Dataset statistics.}
    \label{tab:datasets}
    \scriptsize
    \setlength{\tabcolsep}{1pt}
    \begin{tabular}{l|l|c|c|c}
    \hline
    \multicolumn{2}{l|}{Attribute} & WSDREAM-2T~\cite{wsdream} & Small-3T~\cite{PMT_2023_TNSM} & gRPC-4T~\cite{arrqp_tsc_2025} \\ \hline
    \multicolumn{2}{l|}{Users / services}            & 339 / 5,825   & 112 / 36    & 57 / 150 \\
    \multicolumn{2}{l|}{User regions / AS}         & 31 / 137      & 24 / 80     & 13 / 16 \\
    \multicolumn{2}{l|}{Service regions / AS}      & 74 / 2,699    & 16 / 28     & 1 / 1 \\
    \multicolumn{2}{l|}{QoS records (density)}       & 1,831,253 (92.7\%) & 2,870 (71.2\%) & 8,550 (100\%) \\ \hline
    \multirow{2}{*}{RT (s)}   & (min, max)  & (0.001, 19.990) & (0.011, 25.806) & -- \\
                              & mean $\pm$ std & 0.909 $\pm$ 1.973 & 1.267 $\pm$ 2.523 & -- \\ \hline
    \multirow{2}{*}{TP (kbps)}& (min, max)  & (0.004, 1000.000) & (0.648, 3454.468) & -- \\
                              & mean $\pm$ std & 47.562 $\pm$ 110.797 & 35.090 $\pm$ 132.533 & -- \\ \hline
    \multirow{2}{*}{RE}       & (min, max)  & -- & (0.010, 1.000) & (0.446, 1.000) \\
                              & mean $\pm$ std & -- & 0.990 $\pm$ 0.059 & 0.684 $\pm$ 0.113 \\ \hline
    \multirow{2}{*}{CT} & (min, max)  & -- & -- & (4.600, 17.600) \\
                              & mean $\pm$ std & -- & -- & 11.221 $\pm$ 1.954 \\ \hline
    \multirow{2}{*}{LT (s)}  & (min, max)  & -- & -- & (0.048, 151.864) \\
                              & mean $\pm$ std & -- & -- & 3.779 $\pm$ 9.522 \\ \hline
    \multirow{2}{*}{PW (W)} & (min, max)  & -- & -- & (17.210, 92.750) \\
                              & mean $\pm$ std & -- & -- & 56.504 $\pm$ 20.000 \\ \hline
    \end{tabular}
\end{table}

\begin{table*}[!t]
\centering
\caption{Comparison with joint QoS prediction methods on WSDREAM-2T, Small-3T, and gRPC-4T under four training densities (TD) on MAE and RMSE error metrics.}
\label{tab:mtl_sota_comparison}
\scriptsize
\setlength{\tabcolsep}{2pt}
\renewcommand{\arraystretch}{0.87}
\begin{tabular*}{0.8\textwidth}{@{\extracolsep{\fill}}cl cccc cccc@{}}
\hline
\multirow{2}{*}{QoS} & \multirow{2}{*}{Method} & \multicolumn{4}{c}{MAE (TD, $\%$) $\downarrow$}  & \multicolumn{4}{c}{RMSE (TD, $\%$) $\downarrow$} \\ \cline{3-6}\cline{7-10}
 & & 5 & 10 & 15 & 20 & 5 & 10 & 15 & 20 \\ \hline
\multicolumn{10}{c}{\cellcolor{gray!15}\textbf{WSDREAM-2T}} \\ \hline
 \multirow{7}{*}{RT} & JQSP & 0.5079 & 0.4406 & 0.4154 & 0.4072 & 1.9491 & 1.4445 & 1.3496 & 1.3181 \\
  & WAMTL & 0.4652 & 0.4097 & 0.3844 & 0.3663 & 1.3607 & 1.2800 & 1.2345 & 1.2017 \\
  & PLHEN & 0.4449 & 0.3863 & 0.3630 & 0.3449 & 1.3623 & 1.2691 & 1.2202 & \underline{1.1878} \\
  & DNM & 0.4121 & 0.3621 & 0.3471 & 0.3214 & 1.3866 & 1.2670 & 1.2216 & 1.2070 \\
  & MGEN & \underline{0.4115} & \underline{0.3423} & \underline{0.3289} & \underline{0.3181} & \underline{1.3579} & \underline{1.2602} & \underline{1.2189} & 1.1950 \\ \cline{2-10}
  & \textbf{SHARP-QoS} & \textbf{0.3668} & \textbf{0.3234} & \textbf{0.3039} & \textbf{0.2894} & \textbf{1.3260} & \textbf{1.2496} & \textbf{1.2076} & \textbf{1.1694} \\
  & $I$ (\%) & 10.86 & 5.52 & 7.60 & 9.02 & 2.35 & 0.84 & 0.93 & 1.55 \\
\hline
 \multirow{7}{*}{TP} & JQSP & 21.9919 & 17.9796 & 16.8000 & 14.3989 & 79.5323 & 54.0659 & 49.8281 & 49.0998 \\
  & WAMTL & 18.8521 & 15.3414 & 14.5306 & 13.7570 & 54.5032 & 46.7399 & 43.5842 & 41.5083 \\
  & PLHEN & \underline{14.4471} & \underline{12.3166} & \underline{11.2453} & \underline{10.9607} & \underline{47.9303} & \underline{41.6356} & \underline{39.7921} & \underline{37.2607} \\
  & DNM & 17.2980 & 14.0650 & 14.5436 & 14.1170 & 58.1629 & 50.3930 & 49.8868 & 46.9100 \\
  & MGEN & 15.4529 & 13.0833 & 12.6516 & 12.3941 & 50.7213 & 43.3581 & 43.2147 & 42.1618 \\ \cline{2-10}
  & \textbf{SHARP-QoS} & \textbf{12.8962} & \textbf{11.0828} & \textbf{10.4246} & \textbf{9.8934} & \textbf{46.2917} & \textbf{40.7926} & \textbf{39.0727} & \textbf{36.5829} \\
  & $I$ (\%) & 10.74 & 10.02 & 7.30 & 9.74 & 3.42 & 2.02 & 1.81 & 1.82 \\
\hline
\multicolumn{10}{c}{\cellcolor{gray!15}\textbf{Small-3T}} \\ \hline
 \multirow{7}{*}{RT} & JQSP & 0.7688 & 0.5053 & 0.4495 & 0.4650 & 1.6266 & 1.3327 & \underline{1.0606} & 1.1111 \\
  & WAMTL & \underline{0.6562} & 0.4771 & 0.4423 & 0.4093 & \underline{1.2595} & 1.2268 & 1.1525 & 1.1269 \\
  & PLHEN & 0.7804 & 0.5697 & 0.5046 & 0.4924 & 2.2360 & 1.8068 & 1.7642 & 1.6837 \\
  & DNM & 1.1098 & 1.0413 & 1.0332 & 1.0016 & 2.4307 & 2.1130 & 2.0288 & 2.0040 \\
  & MGEN & 0.6627 & \underline{0.4216} & \underline{0.3794} & \underline{0.3639} & 1.4639 & \underline{1.1895} & 1.0862 & \underline{1.0790} \\ \cline{2-10}
  & \textbf{SHARP-QoS} & \textbf{0.4496} & \textbf{0.3945} & \textbf{0.3099} & \textbf{0.2871} & \textbf{1.1652} & \textbf{1.1015} & \textbf{1.0263} & \textbf{1.0029} \\
  & $I$ (\%) & 31.48 & 6.43 & 18.32 & 21.10 & 7.49 & 7.40 & 3.23 & 7.05 \\
\hline
 \multirow{7}{*}{TP} & JQSP & 54.2766 & 46.9502 & 50.9577 & 27.0942 & 253.2556 & 218.3714 & 112.5865 & 92.1144 \\
  & WAMTL & 33.6497 & 27.6057 & 23.2590 & 18.9285 & 132.5707 & 129.2284 & 126.6743 & 111.6510 \\
  & PLHEN & 26.2481 & 22.0969 & 18.4481 & 17.7547 & 129.0895 & 112.8360 & 105.5944 & 100.5029 \\
  & DNM & 43.2913 & 33.4560 & 35.6610 & 34.6610 & \underline{116.1662} & 109.6389 & 106.0478 & 102.7562 \\
  & MGEN & \underline{24.5219} & \underline{18.1270} & \underline{15.0493} & \underline{14.9293} & 137.2764 & \underline{107.0232} & \underline{99.8948} & \underline{78.2624} \\ \cline{2-10}
  & \textbf{SHARP-QoS} & \textbf{24.4732} & \textbf{17.7958} & \textbf{14.4464} & \textbf{11.4730} & \textbf{88.3392} & \textbf{71.9799} & \textbf{72.4821} & \textbf{43.4623} \\
  & $I$ (\%) & 0.20 & 1.83 & 4.01 & 23.15 & 23.95 & 32.74 & 27.44 & 44.47 \\
\hline
 \multirow{7}{*}{RE} & JQSP & 0.0429 & 0.0348 & 0.0181 & 0.0116 & 0.0776 & 0.0662 & \underline{0.0455} & \underline{0.0426} \\
  & WAMTL & 0.2805 & 0.0941 & 0.0907 & 0.0699 & 0.3528 & 0.1383 & 0.1148 & 0.1094 \\
  & PLHEN & 0.0822 & 0.0276 & 0.0237 & 0.0272 & 0.1994 & 0.0896 & 0.0862 & 0.0834 \\
  & DNM & \underline{0.0112} & \underline{0.0109} & \underline{0.0101} & \underline{0.0091} & \underline{0.0572} & \underline{0.0547} & 0.0484 & 0.0459 \\
  & MGEN & 0.0484 & 0.0395 & 0.0164 & 0.0158 & 0.1926 & 0.1830 & 0.0972 & 0.0964 \\ \cline{2-10}
  & \textbf{SHARP-QoS} & \textbf{0.0100} & \textbf{0.0072} & \textbf{0.0062} & \textbf{0.0060} & \textbf{0.0344} & \textbf{0.0339} & \textbf{0.0326} & \textbf{0.0299} \\
  & $I$ (\%) & 10.71 & 33.94 & 38.61 & 34.07 & 39.86 & 38.03 & 28.35 & 29.81 \\
\hline
\multicolumn{10}{c}{\cellcolor{gray!15}\textbf{gRPC-4T}} \\ \hline
 \multirow{7}{*}{RE} & JQSP & 0.0879 & 0.0753 & 0.0654 & 0.0626 & 0.1545 & 0.1329 & 0.1226 & 0.1139 \\
  & WAMTL & 0.0475 & 0.0360 & 0.0290 & 0.0240 & 0.0688 & 0.0498 & 0.0423 & 0.0325 \\
  & PLHEN & \underline{0.0329} & \underline{0.0283} & \underline{0.0174} & \underline{0.0166} & \underline{0.0498} & \underline{0.0491} & \underline{0.0421} & \underline{0.0308} \\
  & DNM & 0.0505 & 0.0345 & 0.0328 & 0.0310 & 0.0663 & 0.0495 & 0.0476 & 0.0449 \\
  & MGEN & 0.0422 & 0.0287 & 0.0253 & 0.0233 & 0.0706 & 0.0564 & 0.0440 & 0.0413 \\ \cline{2-10}
  & \textbf{SHARP-QoS} & \textbf{0.0295} & \textbf{0.0200} & \textbf{0.0160} & \textbf{0.0123} & \textbf{0.0443} & \textbf{0.0475} & \textbf{0.0416} & \textbf{0.0286} \\
  & $I$ (\%) & 10.33 & 29.33 & 8.05 & 25.90 & 11.04 & 3.26 & 1.19 & 7.14 \\
\hline
 \multirow{7}{*}{CT} & JQSP & \underline{1.0381} & \underline{0.8232} & 0.7852 & 0.7807 & \underline{1.3539} & \underline{1.0784} & 1.0124 & 0.9981 \\
  & WAMTL & 1.1504 & 0.9515 & 0.7851 & \underline{0.6334} & 1.4946 & 1.2406 & 1.0152 & 1.0023 \\
  & PLHEN & 1.2472 & 1.1667 & 1.1416 & 1.1271 & 1.7155 & 1.6308 & 1.5848 & 1.5596 \\
  & DNM & 1.1464 & 0.8296 & \underline{0.6902} & 0.6659 & 1.5566 & 1.1257 & \underline{0.9239} & \underline{0.8922} \\
  & MGEN & 1.5543 & 1.1733 & 0.8303 & 0.7219 & 2.2099 & 1.5409 & 1.1259 & 0.9951 \\ \cline{2-10}
  & \textbf{SHARP-QoS} & \textbf{0.8836} & \textbf{0.7060} & \textbf{0.6008} & \textbf{0.5618} & \textbf{1.2044} & \textbf{0.9976} & \textbf{0.8452} & \textbf{0.8020} \\
  & $I$ (\%) & 14.88 & 14.24 & 12.95 & 11.30 & 11.04 & 7.49 & 8.52 & 10.11 \\
\hline
 \multirow{7}{*}{LT} & JQSP & 4.8826 & 3.9492 & 3.6310 & 3.3331 & 9.4759 & \underline{7.7057} & \underline{7.5593} & 7.4431 \\
  & WAMTL & 3.6652 & 3.5620 & 3.4255 & 2.5137 & 9.2406 & 8.2041 & 8.0149 & 7.9424 \\
  & PLHEN & 3.9556 & 3.6657 & 3.5486 & 3.4720 & 9.3877 & 8.7526 & 8.6382 & 8.5536 \\
  & DNM & 4.3804 & 4.2163 & 3.4580 & 3.4164 & 9.2961 & 8.3004 & 7.9792 & 7.9022 \\
  & MGEN & \underline{2.6290} & \underline{2.0706} & \underline{2.0584} & \underline{2.0523} & \underline{9.0301} & 8.6840 & 8.4091 & \underline{5.6198} \\ \cline{2-10}
  & \textbf{SHARP-QoS} & \textbf{1.1910} & \textbf{0.8909} & \textbf{0.8241} & \textbf{0.7896} & \textbf{1.9780} & \textbf{1.7863} & \textbf{1.5810} & \textbf{1.5799} \\
  & $I$ (\%) & 54.70 & 56.97 & 59.96 & 61.53 & 78.10 & 76.82 & 79.09 & 71.89 \\
\hline
 \multirow{7}{*}{PW} & JQSP & 13.8065 & 10.5663 & 10.0520 & 9.7656 & 17.7371 & \underline{13.9167} & 13.2140 & 12.6584 \\
  & WAMTL & 12.8474 & 9.2137 & 8.4201 & 7.3705 & 17.8271 & 16.3605 & 14.3976 & 12.9835 \\
  & PLHEN & \underline{12.1504} & \underline{8.0194} & 7.1880 & 7.0331 & \underline{17.0120} & 14.1469 & 12.3071 & 11.9526 \\
  & DNM & 12.2556 & 9.0397 & 8.5686 & 8.2758 & 17.3730 & 14.3611 & 12.5894 & 12.1153 \\
  & MGEN & 12.4266 & 8.0331 & \underline{6.2375} & \underline{5.6198} & 19.7486 & 14.4056 & \underline{12.0813} & \underline{11.1539} \\ \cline{2-10}
  & \textbf{SHARP-QoS} & \textbf{11.3154} & \textbf{7.2178} & \textbf{6.2153} & \textbf{5.0122} & \textbf{16.8738} & \textbf{13.6852} & \textbf{12.0119} & \textbf{10.7687} \\
  & $I$ (\%) & 6.87 & 10.00 & 0.36 & 10.81 & 0.81 & 1.66 & 0.57 & 3.45 \\
\hline
\end{tabular*}

\smallskip
\parbox{0.8\textwidth}{\scriptsize
Best results are in \textbf{bold}; the strongest baseline is \underline{underlined}. $I$ (\%) denotes SHARP-QoS's relative improvement over the strongest baseline.
}
\end{table*}

\subsection{Performance Comparison}\label{sec:result_analysis}
We compare SHARP-QoS with five joint QoS prediction methods: DNM~\cite{DNM_2021_TSC}, JQSP~\cite{JQSP_2023_TNSM}, MGEN~\cite{MGEN_2023_JP}, WAMTL~\cite{WAMTL_2024_ICWS}, and PLHEN~\cite{PLHEN_ESA_2026}, using MAE, RMSE, and computational efficiency. 
HTG~\cite{HTT_2024_ETT} and PMT~\cite{PMT_2023_TNSM} are excluded because their required AS-level topology links are not publicly available.

\textbf{(i) Comparison with joint QoS prediction methods.}
Table~\ref{tab:mtl_sota_comparison} compares the methods on three datasets at training densities of $5\%$--$20\%$. We make the following observations.

\textit{Performance across datasets and QoS attributes.}
SHARP-QoS achieves the lowest MAE and RMSE in all settings, although the gains vary across datasets and QoS attributes. On WSDREAM-2T, improvements are modest, with RT RMSE gains below $3\%$. On Small-3T, the gains reach $44.47\%$ for TP and $39.86\%$ for RE in RMSE. On gRPC-4T, LT shows the largest gains, with MAE and RMSE reductions of up to $61.53\%$ and $79.09\%$, respectively. Some improvements are marginal, such as the $0.20\%$ MAE reduction for Small-3T TP at $5\%$ density. These results show that the gains depend on the dataset and QoS attribute.

\textit{Performance with sparse observations.}
SHARP-QoS outperforms all baselines even at $5\%$ training density, reducing MAE by $31.48\%$ for Small-3T RT and $54.70\%$ for gRPC-4T LT; the corresponding LT RMSE reduction is $78.10\%$. Errors generally decrease as training density increases, while SHARP-QoS retains its advantage across all densities.

\textit{Reduction in large errors.}
For several attributes, RMSE improves more than MAE, particularly for Small-3T TP and gRPC-4T LT. Since RMSE penalizes large errors more heavily, this indicates that SHARP-QoS also reduces larger prediction errors. The heavy-tailed distributions of both attributes (Table~\ref{tab:datasets}) may partly explain these gains. In contrast, improvements in WSDREAM-2T RT RMSE remain modest.

\textit{Comparison among baselines.}
No baseline consistently ranks second across all settings. MGEN performs strongly on WSDREAM-2T RT, Small-3T TP (MAE), and gRPC-4T LT; PLHEN on WSDREAM-2T TP and gRPC-4T RE; DNM on Small-3T RE; and WAMTL in some Small-3T RT settings. SHARP-QoS, however, achieves the lowest errors across all datasets, attributes, and training densities.

\begin{table}[!t]
    \centering
    \caption{Computational cost on WSDREAM-2T. Inference time is per prediction.}
    \label{tab:computational_complexity}
    \scriptsize
    \setlength{\tabcolsep}{0pt}
    \begin{tabular*}{\columnwidth}{@{\extracolsep{\fill}}lccc@{}}
        \hline
       {Method} & Params (M) $\downarrow$ & FLOPs (G) $\downarrow$ & Inference ($\times10^{-8}$\,s) $\downarrow$ \\ \hline
        JQSP~\cite{JQSP_2023_TNSM}   & \underline{1.02} & 3207.91 & 79.40 \\
        WAMTL~\cite{WAMTL_2024_ICWS} & 2.96 & 6863.40 & 50.20 \\
        PLHEN~\cite{PLHEN_ESA_2026}  & 16.56 & 64076.28 & 508.00 \\
        DNM~\cite{DNM_2021_TSC}      & \textbf{0.56} & \textbf{90.49} & \textbf{4.81} \\
        MGEN~\cite{MGEN_2023_JP}     & 1.98 & 5872.59 & 66.30 \\ \hline
        SHARP-QoS                    & 1.31 & \underline{385.91} & \underline{7.08} \\ \hline
    \end{tabular*}
    
    \smallskip
    \parbox{\textwidth}{\scriptsize
    Best results are in \textbf{bold}; the strongest baseline is \underline{underlined}.
    }
\end{table}

\textbf{(ii) Computational cost.}
Table~\ref{tab:computational_complexity} compares the computational cost on WSDREAM-2T. SHARP-QoS requires 385.91\,G FLOPs, fewer than every baseline except DNM, and 1.31\,M parameters. The contrast is sharpest with the expert-based methods: PLHEN uses about $12.6\times$ more parameters and $166\times$ more FLOPs, while MGEN uses about $15\times$ more FLOPs. These results indicate that routing-based sharing can reduce model complexity compared with the evaluated expert-based architectures. SHARP-QoS also achieves the second-lowest inference time per prediction ($7.08\times10^{-8}$\,s), after DNM, making it suitable for latency-sensitive QoS prediction.

The additional computational costs associated with training and GPU memory consumption are examined separately in Appendix~A.


\subsection{Ablation Study}
This section examines the contributions of the hierarchical feature extraction backbone, individual modules, fusion mechanism, and graph configuration.

\begin{table}[!t]
\centering
\caption{Graph convolution backbones for the QoS graphs on WSDREAM-2T.}
\label{tab:hgconv_ablation}
\scriptsize
\setlength{\tabcolsep}{0pt}
\begin{tabular*}{\columnwidth}{@{\extracolsep{\fill}}l cccc cccc@{}}
\hline
\multirow{3}{*}{Backbone} & \multicolumn{4}{c}{TD = 10\%} & \multicolumn{4}{c}{TD = 20\%} \\ \cline{2-5}\cline{6-9}
 & \multicolumn{2}{c}{RT} & \multicolumn{2}{c}{TP} & \multicolumn{2}{c}{RT} & \multicolumn{2}{c}{TP} \\
 & MAE & RMSE & MAE & RMSE & MAE & RMSE & MAE & RMSE \\ \hline
GCN & 0.3291 & 1.2628 & 11.2514 & 41.2364 & 0.2941 & 1.1807 & 10.0092 & 37.5613 \\
HGCN & 0.3278 & 1.2570 & 11.2415 & 41.1169 & 0.2936 & 1.1800 & 10.0202 & 37.2023 \\
HyGCN & \underline{0.3254} & \underline{1.2511} & \underline{11.1710} & \underline{40.9829} & \underline{0.2909} & \underline{1.1749} & \underline{9.9972} & \underline{36.9759} \\ \hline
HHGCN (ours)                & \textbf{0.3234} & \textbf{1.2496} & \textbf{11.0828} & \textbf{40.7926} & \textbf{0.2894} & \textbf{1.1694} & \textbf{9.8934} & \textbf{36.5829} \\ \hline
\end{tabular*}

\smallskip
    \parbox{\textwidth}{\scriptsize
    Best results are in \textbf{bold}; the strongest baseline is \underline{underlined}.
    }
\end{table}

\textbf{(i) Hierarchical feature extraction.}
Table~\ref{tab:hgconv_ablation} compares HHGCN with three alternatives for the QoS graphs: Euclidean graph convolution (GCN), Euclidean hypergraph convolution (HGCN), and hyperbolic graph convolution without hyperedges (HyGCN). HHGCN achieves the lowest MAE and RMSE for both RT and TP at both densities. Compared with GCN, it reduces MAE by $1.73\%$ for RT and $1.50\%$ for TP at $10\%$ density, and the largest RMSE reduction is $2.60\%$ for TP at $20\%$. Most of this gain comes from the hyperbolic geometry: moving from GCN to HyGCN already reduces RT MAE by about $1.1\%$, whereas adding hyperedges in Euclidean space (HGCN) helps little and slightly increases TP MAE at $20\%$. Combined with hyperbolic convolution, however, the hyperedges reduce both MAE and RMSE further over HyGCN, so the two components are complementary. Since the region and AS graphs have no hyperedges, we use HyGCN for them throughout.

\begin{table}[!t]
\centering
\caption{Module ablation and fusion variants on WSDREAM-2T. ``w/o'' removes HHGCN (H), contextual SNR (S), or Cross-SNR (C); fusion variants replace gated fusion.}
\label{tab:module_ablation}
\scriptsize
\setlength{\tabcolsep}{0pt}
\begin{tabular*}{\columnwidth}{@{\extracolsep{\fill}}l cccc cccc@{}}
\hline
\multirow{3}{*}{Variant} & \multicolumn{4}{c}{TD = 10\%} & \multicolumn{4}{c}{TD = 20\%} \\ \cline{2-5}\cline{6-9}
 & \multicolumn{2}{c}{RT} & \multicolumn{2}{c}{TP} & \multicolumn{2}{c}{RT} & \multicolumn{2}{c}{TP} \\
 & MAE & RMSE & MAE & RMSE & MAE & RMSE & MAE & RMSE \\ \hline
\multicolumn{9}{l}{\cellcolor{gray!15}\textit{Module removal}} \\
w/o C    & 0.3278 & 1.2585 & {11.1198} & {41.0943} & 0.2938 & 1.2040 & {10.0173} & 37.1704 \\
w/o S    & \underline{0.3241} & {1.2531} & 11.1678 & 41.2077 & 0.2924 & 1.1790 & 10.0208 & 37.7488 \\
w/o H    & 0.3267 & 1.2570 & 11.4262 & 41.4448 & 0.2955 & 1.1796 & 10.0867 & 37.2375 \\
w/o H, S & 0.3294 & 1.2600 & 11.4375 & 41.5981 & 0.2949 & \underline{1.1738} & 10.1076 & 37.6458 \\
w/o S, C & 0.3265 & 1.2657 & 11.2696 & 41.1782 & 0.2998 & 1.2082 & 10.0573 & 37.2478 \\
w/o H, C & 0.3448 & 1.2926 & 11.7652 & 41.8285 & 0.3112 & 1.2092 & 10.6097 & 38.5355 \\
\multicolumn{9}{l}{\cellcolor{gray!15}\textit{Fusion alternatives}} \\
Concatenation & 0.3242 & \underline{1.2507} & 11.0978 & \underline{40.9620} & 0.2923 & 1.1832 & 10.0657 & \underline{37.0781} \\
Attention     & 0.3276 & 1.2587 & \textbf{11.0693} & 41.2112 & 0.2935 & 1.1862 & \underline{10.0071} & 37.7384 \\
Summation     & 0.3272 & 1.2561 & 11.2413 & 41.1458 & \underline{0.2910} & {1.1770} & 10.0744 & 37.8381 \\ \hline
Gated (full)  & \textbf{0.3234} & \textbf{1.2496} & \underline{11.0828} & \textbf{40.7926} & \textbf{0.2894} & \textbf{1.1694} & \textbf{9.8934} & \textbf{36.5829} \\ \hline
\end{tabular*}

\smallskip
    \parbox{\textwidth}{\scriptsize
    Best results are in \textbf{bold}; the strongest baseline is \underline{underlined}.
    }
\end{table}

\textbf{(ii) Module ablation and fusion.}
Table~\ref{tab:module_ablation} evaluates removing HHGCN (replaced by an MLP on NMF features), contextual SNR, Cross-SNR, or pairs of these modules. At $20\%$ density, all removals increase the four error metrics. Across both densities, single-module removals increase errors by $0.22\%$--$3.19\%$, while removing HHGCN and Cross-SNR together causes the largest degradation ($2.54\%$--$7.53\%$). This highlights the complementary roles of hierarchical structural learning and cross-QoS feature sharing. At $10\%$ density, individual effects are generally smaller, except for a $3.10\%$ increase in TP MAE after removing HHGCN. The larger joint degradation further supports the complementary contributions of structural representations and feature sharing.

{The second part compares gated fusion (\S~\ref{subsec:gated_feat_fusion}) with concatenation, attention, and summation. Gated fusion achieves the lowest error in seven of eight settings, outperforming the best alternative by up to $1.34\%$. The exception is TP MAE at $10\%$ density, where attention performs better by $0.12\%$. Overall, adaptive gating provides more effective feature integration than the evaluated alternatives.}

\textbf{(iii) Graph configuration.}
Table~\ref{tab:context_gconv} compares graph configurations. Using only the QoS graph, or merging the QoS, region, and AS edges into a single graph, yields higher errors than SHARP-QoS; the merged graph is no better than the QoS graph alone, so simply adding contextual edges to one graph does not help. Adding the region or AS graph as a separate input improves on the QoS-only variant, but neither is consistently better: at $10\%$ density, region gives lower RT MAE and RMSE and TP MAE, while AS gives lower TP RMSE; at $20\%$, AS gives lower RT MAE and RMSE and TP MAE, while region gives lower TP RMSE. Learning from all three graphs separately achieves the lowest MAE and RMSE in every setting, which suggests that invocation patterns, shared network (AS) membership, and geographic proximity carry complementary information that is better kept in separate graphs than merged into one.

\begin{table}[!t]
\centering
\caption{Graph configurations on WSDREAM-2T. ``Merged'' combines QoS, region, and AS edges in one graph; SHARP-QoS (Ours) learns from each graph separately.}
\label{tab:context_gconv}
\scriptsize
\setlength{\tabcolsep}{0pt}
\begin{tabular*}{\columnwidth}{@{\extracolsep{\fill}}l cccc cccc@{}}
\hline
\multirow{3}{*}{Graphs} & \multicolumn{4}{c}{TD = 10\%} & \multicolumn{4}{c}{TD = 20\%} \\ \cline{2-5}\cline{6-9}
 & \multicolumn{2}{c}{RT} & \multicolumn{2}{c}{TP} & \multicolumn{2}{c}{RT} & \multicolumn{2}{c}{TP} \\
 & MAE & RMSE & MAE & RMSE & MAE & RMSE & MAE & RMSE \\ \hline
QoS only                 & 0.3317 & 1.2734 & 11.1948 & 41.3736 & 0.2949 & 1.1871 & 10.1523 & 37.2919 \\
QoS + region                 & \underline{0.3285} & \underline{1.2596} & \underline{11.1704} & 41.3532 & 0.2915 & 1.1752 & 9.9939 & \underline{37.2060} \\
QoS + AS                 & 0.3298 & 1.2676 & 11.1856 & \underline{41.1025} & \underline{0.2907} & \underline{1.1721} & \underline{9.9293} & 37.2759 \\
Merged & 0.3299 & 1.2732 & 11.5039 & 42.0504 & 0.2961 & 1.1994 & 10.0540 & 37.4746 \\ \hline
Ours & \textbf{0.3234} & \textbf{1.2496} & \textbf{11.0828} & \textbf{40.7926} & \textbf{0.2894} & \textbf{1.1694} & \textbf{9.8934} & \textbf{36.5829} \\ \hline
\end{tabular*}

\smallskip
    \parbox{\textwidth}{\scriptsize
    Best results are in \textbf{bold}; the strongest baseline is \underline{underlined}.
    }
\end{table}


\pgfplotsset{altsingle/.style={
    width=\dimexpr\linewidth-13pt\relax,
    height=3.8cm,
    ybar=1pt,
    bar width=5pt,
    ymin=0,
    xmin=0.4,
    xmax=6.6,
    xtick={1,2,3,4,5,6},
    xticklabels={PMF,CMF,DCALF,llmQoS,QoSGNN,\textbf{SHARP-QoS}},
    axis lines*=left,
    axis line style={black!40},
    tick style={black!40},
    axis on top,
    ymajorgrids=true,
    grid style={black!10},
    scaled y ticks=false,
    yticklabel style={/pgf/number format/fixed},
    tick label style={font=\scriptsize},
    xticklabel style={rotate=35,anchor=east,font=\scriptsize},
    ylabel style={font=\scriptsize},
    legend style={
        at={(0.5,1.03)},
        anchor=south,
        draw=none,
        fill=none,
        font=\scriptsize,
        legend columns=2
    },
    legend cell align=left
}}

\begin{figure}[!t]
\centering

\begin{subfigure}[t]{0.49\textwidth}
\centering
\begin{tikzpicture}
\begin{axis}[
    altsingle,
    ylabel={RT MAE},
    ymax=0.55,
    ytick={0,0.1,0.2,0.3,0.4,0.5}
]
\path[fill=altMixHighlight]
    (axis cs:5.5,0) rectangle (axis cs:6.5,0.55);

\addplot[
    fill=altMixTen,
    draw=altMixTen!80!black,
    line width=0.25pt
] coordinates {
    (1,0.4996)
    (2,0.4511)
    (3,0.4544)
    (4,0.3600)
    (5,0.3450)
    (6,0.3234)
};
\addlegendentry{Density $10\%$}

\addplot[
    fill=altMixTwenty,
    draw=altMixTwenty!80!black,
    line width=0.25pt
] coordinates {
    (1,0.4492)
    (2,0.3767)
    (3,0.4246)
    (4,0.3270)
    (5,0.3044)
    (6,0.2894)
};
\addlegendentry{Density $20\%$}

\end{axis}
\end{tikzpicture}
\vspace{-10pt}
\caption{RT prediction.}
\end{subfigure}
\hfill
\begin{subfigure}[t]{0.49\textwidth}
\centering
\begin{tikzpicture}
\begin{axis}[
    altsingle,
    ylabel={TP MAE},
    ymax=26,
    ytick={0,5,10,15,20,25}
]
\path[fill=altMixHighlight]
    (axis cs:5.5,0) rectangle (axis cs:6.5,26);

\addplot[
    fill=altMixTen,
    draw=altMixTen!80!black,
    line width=0.25pt
] coordinates {
    (1,16.1755)
    (2,23.2347)
    (3,15.3595)
    (4,12.0220)
    (5,13.9460)
    (6,11.0828)
};

\addplot[
    fill=altMixTwenty,
    draw=altMixTwenty!80!black,
    line width=0.25pt
] coordinates {
    (1,14.6694)
    (2,18.8050)
    (3,13.6697)
    (4,10.7600)
    (5,12.1049)
    (6,9.8934)
};

\end{axis}
\end{tikzpicture}
\vspace{-10pt}
\caption{TP prediction.}
\end{subfigure}

\caption{Single-QoS methods versus SHARP-QoS on WSDREAM-2T at $10\%$ and $20\%$ training densities. Shading highlights SHARP-QoS; lower MAE is better.}
\label{fig:alt_single_task}

\end{figure}

\subsection{Comparison with Alternative Prediction and Optimization Methods}

We evaluate SHARP-QoS against single-QoS predictors, classical MTL architectures, architectural replacements, and alternative task-balancing strategies on WSDREAM-2T.

\textbf{(i) Comparison with single-QoS prediction methods.}
Fig.~\ref{fig:alt_single_task} compares SHARP-QoS with five single-QoS predictors: PMF~\cite{pmf_nips_2007}, CMF~\cite{cmf_www21}, DCALF~\cite{dcalf_tkde_2022}, llmQoS~\cite{llmQoS_2025_sse}, and QoSGNN~\cite{QoSGNN_TSC_2024}. SHARP-QoS achieves the lowest MAE for both tasks at $10\%$ and $20\%$ training densities. Compared with QoSGNN, the strongest RT baseline, it reduces RT MAE by $6.26\%$ and $4.93\%$, respectively. Against llmQoS, the strongest TP baseline, the corresponding TP MAE reductions are $7.81\%$ and $8.05\%$. Thus, SHARP-QoS improves prediction accuracy for both parameters within a single framework.

{\textbf{(ii) Comparison with classical MTL architectures.}
We compare SHARP-QoS with three MTL architectures build on NMF-based features: Cross-Stitch~\cite{misra2016_cross-stich}, MMoE~\cite{ma2018_MMoE}, and PLE~\cite{tang2020_PLE}. As shown in Table~\ref{tab:alt_mtl_variants}, SHARP-QoS achieves the lowest MAE across both tasks and densities. Relative to the strongest classical MTL baseline in each setting, it reduces RT MAE by $9.00\%$ and $10.49\%$ and TP MAE by $11.23\%$ and $12.12\%$ at $10\%$ and $20\%$ density, respectively.}

{\textbf{(iii) Alternative backbone, sharing mechanisms, and capacity-matched network.}
Table~\ref{tab:alt_mtl_variants} also evaluates architectural replacements within SHARP-QoS. Replacing HHGCN with SGFormer~\cite{sgformer_NEURIPS2023} increases MAE for both tasks at both densities. Similarly, replacing SNR and Cross-SNR with Cross-Stitch~\cite{misra2016_cross-stich}, PLE~\cite{tang2020_PLE}, MMoE~\cite{ma2018_MMoE}, AdaTT~\cite{adatt_kdd_2023}, or DSelect-k~\cite{dselectk_NEURIPS2021} yields higher MAE in every setting. The parameter-matched FFN (PM-FFN) also performs worse, indicating that the improvement is not explained by parameter count alone.}

{\textbf{(iv) Comparison with multi-task optimization strategies.}
Table~\ref{tab:alt_optimization} compares EMA with gradient-based methods (NashMTL~\cite{nashmtl}, CAGrad~\cite{cagrad}, PCGrad~\cite{pcgrad}, and MGDA~\cite{mgda}) and loss-weighting methods (GradNorm~\cite{gradnorm}, FAMO~\cite{famo_NEURIPS2023}, HUW~\cite{Uncertainty_CVPR_2018}, UTW, and DWA~\cite{dwa_2019_CVPR}). EMA achieves the lowest MAE in three of four settings, including TP at both densities and RT at $20\%$ density. DWA marginally outperforms EMA on RT at $10\%$ density ($0.3233$ versus $0.3234$). However, EMA reduces TP MAE relative to DWA by $5.78\%$ and $3.31\%$ at $10\%$ and $20\%$ density, respectively. Compared with UTW, EMA reduces RT and TP MAE by $1.22\%$ and $6.45\%$ at $10\%$ density, and $2.33\%$ and $3.48\%$ at $20\%$ density.}

\begin{table}[!t]
    \centering
    \caption{Classical MTL architectures and SHARP-QoS variants on WSDREAM-2T (MAE).}
    \label{tab:alt_mtl_variants}
    \scriptsize
    \setlength{\tabcolsep}{0pt}
    \begin{tabular*}{0.95\columnwidth}{@{\extracolsep{\fill}}lcccc@{}}
        \hline
        Method & \multicolumn{2}{c}{Density $10\%$}
               & \multicolumn{2}{c}{Density $20\%$} \\
        & RT & TP & RT & TP \\ \hline

        \multicolumn{5}{l}{\cellcolor{gray!15}\textit{Classical MTL (NMF features)}} \\
        MMoE~\cite{ma2018_MMoE}
            & 0.3701 & 13.1410 & 0.3395 & 12.7715 \\
        Cross-Stitch~\cite{misra2016_cross-stich}
            & 0.3634 & 12.4842 & 0.3233 & 11.2576 \\
        PLE~\cite{tang2020_PLE}
            & 0.3554 & 13.7637 & 0.3234 & 12.7865 \\ \hline

        \multicolumn{5}{l}{\cellcolor{gray!15}\textit{SHARP-QoS variants}} \\
        w/ SGFormer~\cite{sgformer_NEURIPS2023}
            & 0.3333 & 11.4105 & 0.2970 & 10.1133 \\
        w/ Cross-Stitch~\cite{misra2016_cross-stich}
            & 0.3347 & 11.4832 & 0.3012 & 10.2248 \\
        w/ PLE~\cite{tang2020_PLE}
            & 0.3343 & 11.4576 & 0.2952 & 10.0863 \\
        w/ MMoE~\cite{ma2018_MMoE}
            & 0.3311 & 11.4744 & 0.2943 & 10.1828 \\
        w/ AdaTT~\cite{adatt_kdd_2023}
            & 0.3249 & \underline{11.2859}
            & \underline{0.2940} & \underline{10.0315} \\
        w/ DSelect-k~\cite{dselectk_NEURIPS2021}
            & \underline{0.3239} & 11.4621
            & 0.2949 & 10.1330 \\
        w/ PM-FFN
            & 0.3252 & 11.3371 & 0.2981 & 10.1249 \\ \hline
        \textbf{SHARP-QoS}
            & \textbf{0.3234} & \textbf{11.0828}
            & \textbf{0.2894} & \textbf{9.8934} \\ \hline
    \end{tabular*}
    \par\smallskip\raggedright\scriptsize
    \texttt{w/} denotes a component replacement.
    PM-FFN replaces SNR and Cross-SNR with
    parameter-matched feed-forward networks.
    Best results are in \textbf{bold}; the strongest baseline is \underline{underlined}.
\end{table}

\begin{table}[!t]
    \centering
    \caption{Multi-task optimization strategies within SHARP-QoS on WSDREAM-2T (MAE).}
    \label{tab:alt_optimization}
    \scriptsize
    \setlength{\tabcolsep}{0pt}
    \begin{tabular*}{0.95\columnwidth}{@{\extracolsep{\fill}}lcccc@{}}
        \hline
        Method & \multicolumn{2}{c}{Density $10\%$}
               & \multicolumn{2}{c}{Density $20\%$} \\
        & RT & TP & RT & TP \\ \hline

        \multicolumn{5}{l}{\cellcolor{gray!15}\textit{Gradient-based}} \\
        NashMTL~\cite{nashmtl}
            & 0.3935 & 36.8322 & 0.3637 & 36.5003 \\
        CAGrad~\cite{cagrad}
            & 0.3666 & 23.0353 & 0.3172 & 13.2636 \\
        PCGrad~\cite{pcgrad}
            & 0.3461 & 11.6027 & 0.3334 & 10.7505 \\
        MGDA~\cite{mgda}
            & 0.3441 & 24.0792 & 0.3060 & 21.2061 \\ \hline

        \multicolumn{5}{l}{\cellcolor{gray!15}\textit{Loss weighting}} \\
        GradNorm~\cite{gradnorm}
            & 0.3388 & \underline{11.3259}
            & 0.2982 & \underline{9.9997} \\
        FAMO~\cite{famo_NEURIPS2023}
            & 0.3322 & 11.3620 & 0.2975 & 10.0775 \\
        HUW~\cite{Uncertainty_CVPR_2018}
            & 0.3285 & 12.0177 & 0.2965 & 10.1501 \\
        UTW
            & 0.3274 & 11.8468 & 0.2963 & 10.2504 \\
        DWA~\cite{dwa_2019_CVPR}
            & \textbf{0.3233} & 11.7625
            & \underline{0.2953} & 10.2319 \\ \hline
        \textbf{EMA (ours)}
            & \underline{0.3234} & \textbf{11.0828}
            & \textbf{0.2894} & \textbf{9.8934} \\ \hline
    \end{tabular*}
    \par\smallskip\raggedright\scriptsize
    UTW denotes uniform task weighting (equal task weights).
    Best results are in \textbf{bold}; the strongest baseline is \underline{underlined}.
\end{table}

{Appendices~C and D report hyperparameter sensitivity and scalability, respectively; Appendices~E--G assess WSDREAM-2T's hierarchy using Gromov $\delta$-hyperbolicity, Ollivier--Ricci curvature, and embedding distortion.}

We next evaluate SHARP-QoS under anomalous QoS values (\S.~\ref{sec:outlier}), cold-start (\S.~\ref{sec:coldstart}), and statistical reliability (\S.~\ref{sec:significance_test}).


\subsection{Outlier Sensitivity Analysis}\label{sec:outlier}
QoS records contain anomalous values caused by network instability, and such outliers can strongly affect prediction~\cite{cmf_www21}. SHARP-QoS therefore trains with the MAE ($L_1$) loss, which is less sensitive to extreme values than the squared ($L_2$) loss.

To measure how much of the remaining error comes from such values, we identify anomalous test records using Isolation Forest~\cite{iforest} and recompute errors after removing the top $2\%$--$10\%$ most anomalous records in $2\%$ increments at each training density. Fig.~\ref{fig:outlier_rt_tp} shows a sharp error reduction after removing the first $2\%$, followed by slower declines. Removing $10\%$ of the test records reduces errors by $45\%$--$66\%$ across metrics and densities ($55.94\%$ on average), with larger reductions for RT than TP and RMSE than MAE. These results indicate that a small fraction of extreme observations accounts for a substantial share of the remaining error, while errors on typical QoS values remain relatively low.

\begin{figure}[!t]
    \centering
    \begin{tikzpicture}
        \begin{groupplot}[
            grid=major,
            group style={group size=2 by 2, horizontal sep=1.2cm, vertical sep=0.8cm},
            width=0.5\columnwidth, height=0.4\columnwidth,
            xmin=-0.5, xmax=10.5, xtick={0,2,4,6,8,10},
            tick label style={font=\scriptsize}, label style={font=\scriptsize},
            title style={font=\scriptsize, yshift=-1.5ex},
            ymajorgrids, grid style={dashed,gray!35},
            every axis plot/.append style={thick, mark size=1.3pt},
            cycle list={{blue!80!black,mark=o},{red!75!black,mark=square},{green!50!black,mark=triangle},{orange!90!black,mark=diamond}},
        ]
        \nextgroupplot[title={(a) RT -- MAE}, ylabel={MAE},
            legend to name=outlierlegend, legend columns=4,
            legend style={font=\scriptsize, draw=none, column sep=4pt}]
        \addplot coordinates {(0,0.3668)(2,0.2657)(4,0.2171)(6,0.1878)(8,0.1646)(10,0.1409)};
        \addplot coordinates {(0,0.3234)(2,0.2282)(4,0.1839)(6,0.1565)(8,0.1367)(10,0.1193)};
        \addplot coordinates {(0,0.3039)(2,0.2135)(4,0.1715)(6,0.1447)(8,0.1263)(10,0.1111)};
        \addplot coordinates {(0,0.2894)(2,0.2057)(4,0.1670)(6,0.1414)(8,0.1237)(10,0.1097)};
        \legend{TD = $5\%$, TD = $10\%$, TD = $15\%$, TD = $20\%$} 
        \nextgroupplot[title={(b) RT -- RMSE}, ylabel={RMSE}]
        \addplot coordinates {(0,1.3260)(2,0.8660)(4,0.7083)(6,0.6375)(8,0.5755)(10,0.4887)};
        \addplot coordinates {(0,1.2496)(2,0.7836)(4,0.6273)(6,0.5571)(8,0.5025)(10,0.4341)};
        \addplot coordinates {(0,1.2076)(2,0.7512)(4,0.5998)(6,0.5286)(8,0.4773)(10,0.4168)};
        \addplot coordinates {(0,1.1694)(2,0.7369)(4,0.5971)(6,0.5287)(8,0.4788)(10,0.4245)};
        \nextgroupplot[title={(c) TP -- MAE}, ylabel={MAE}, xlabel={Test outliers removed (\%)}]
        \addplot coordinates {(0,12.8962)(2,10.2342)(4,8.8112)(6,7.9153)(8,7.3286)(10,6.8659)};
        \addplot coordinates {(0,11.0828)(2,8.9773)(4,7.7931)(6,7.0162)(8,6.4986)(10,6.0906)};
        \addplot coordinates {(0,10.4246)(2,8.3547)(4,7.2410)(6,6.5180)(8,6.0315)(10,5.6502)};
        \addplot coordinates {(0,9.8934)(2,7.9778)(4,6.9126)(6,6.2380)(8,5.7819)(10,5.4197)}; 
        \nextgroupplot[title={(d) TP -- RMSE}, ylabel={RMSE}, xlabel={Test outliers removed (\%)}]
        \addplot coordinates {(0,46.2917)(2,35.3299)(4,29.9018)(6,26.2612)(8,24.1941)(10,21.8976)};
        \addplot coordinates {(0,40.7926)(2,32.5115)(4,27.8842)(6,24.5358)(8,22.4581)(10,20.3740)};
        \addplot coordinates {(0,39.0727)(2,30.6096)(4,26.2719)(6,23.1402)(8,21.1284)(10,19.1100)};
        \addplot coordinates {(0,36.5829)(2,29.1320)(4,24.8047)(6,21.8723)(8,20.0407)(10,18.1149)};
        \end{groupplot}
        \node[anchor=south] at ($(group c1r1.north)!0.5!(group c2r1.north)+(0,0.25cm)$) {\pgfplotslegendfromname{outlierlegend}};
    \end{tikzpicture}
    \caption{Test error of SHARP-QoS on WSDREAM-2T after removing the $2\%$--$10\%$ most anomalous test records identified by Isolation Forest~\cite{iforest}, for each training density (TD).}
    \label{fig:outlier_rt_tp}
\end{figure}
\subsection{Cold-start Analysis}\label{sec:coldstart}

{The cold-start problem arises when a user or service has no observed QoS records, limiting methods that rely on historical interactions. SHARP-QoS addresses this using region and AS graphs $\mathcal{G}^r$ and $\mathcal{G}^a$ (\S.~\ref{sec:preprocessing}), constructed from public attributes. An entity is \emph{cold} if it has no training records or edges in $\mathcal{G}^p$ and the hypergraphs, but retains edges in $\mathcal{G}^r$ and $\mathcal{G}^a$. A test record is cold if either its user or service is cold, and warm otherwise. Following a DropoutNet-inspired strategy~\cite{DropoutNet_NIPS_2017}, we mask the cross-QoS routing contribution for $15\%$ of warm entities during each training epoch and apply the same masking to cold entities during inference. The contextual and task-specific structural branches remain available to the prediction module.}

{At TD~=~$10\%$, we randomly withhold CSP~=~$10\%$ (34 users, 582 services) or $20\%$ (68 users, 1165 services) of users and services, removing all their training records before constructing NMF features and QoS graphs. The test set remains fixed across CSP settings. Each setting uses three runs with identical withheld entities across methods, and all baselines are retrained accordingly.}

{Table~\ref{tab:cold_start} reports cold-start results for all methods in Table~\ref{tab:mtl_sota_comparison}. Cold records are substantially harder to predict than warm ones across all methods. SHARP-QoS achieves the lowest cold-record MAE and RMSE for both QoS parameters at both CSP levels. Its cold TP MAE (26.86 and 27.81) outperforms MGEN (27.43 and 28.18), while its margin over DNM on RT is smaller, particularly in RMSE. SHARP-QoS also maintains the best warm-record performance, with a smaller increase in warm TP MAE from CSP~=~$10\%$ to $20\%$ (+0.35) than MGEN (+1.02) and PLHEN (+0.99). These results demonstrate that the contextual branch enables effective cold-start prediction while preserving accuracy on warm records.}

\begin{table}[!t]
    \centering
    \caption{Cold-start results on WSDREAM-2T (TD = 10\%), averaged over three runs. CSP is the percentage of users and services withheld; cold records involve a withheld entity.}
    \label{tab:cold_start}
    \scriptsize
    \setlength{\tabcolsep}{0pt}
    \begin{tabular*}{\columnwidth}{@{\extracolsep{\fill}}l cccc cccc@{}}
    \hline
    \multirow{3}{*}{Method} & \multicolumn{4}{c}{CSP = 10\%} & \multicolumn{4}{c}{CSP = 20\%} \\ \cline{2-5}\cline{6-9}
     & \multicolumn{2}{c}{RT} & \multicolumn{2}{c}{TP} & \multicolumn{2}{c}{RT} & \multicolumn{2}{c}{TP} \\
     & MAE & RMSE & MAE & RMSE & MAE & RMSE & MAE & RMSE \\ \hline
    \multicolumn{9}{l}{\cellcolor{gray!15}\textit{Cold records}} \\
    WAMTL~\cite{WAMTL_2024_ICWS} & 0.8828 & 2.1813 & 37.7567 & 98.5129 & 0.8709 & 2.0572 & 39.0988 & 101.1953 \\
    JQSP~\cite{JQSP_2023_TNSM}   & 0.8619 & 2.0832 & 35.7774 & 94.1139 & 0.8901 & 1.9557 & 36.1012 & 99.2150 \\
    PLHEN~\cite{PLHEN_ESA_2026}  & 0.8352 & 2.1440 & 29.4874 & 85.6311 & 0.9549 & 2.2848 & 30.9929 & 89.0028 \\
    MGEN~\cite{MGEN_2023_JP}     & 0.6441 & 1.8703 & \underline{27.4286} & \underline{85.2705} & 0.6145 & 1.8096 & \underline{28.1821} & 88.7418 \\
    DNM~\cite{DNM_2021_TSC}      & \underline{0.6168} & \underline{1.8521} & 28.2831 & 86.3596 & \underline{0.6033} & \underline{1.8088} & 29.7606 & \underline{86.8601} \\
    SHARP-QoS                    & \textbf{0.6125} & \textbf{1.8515} & \textbf{26.8622} & \textbf{80.5579} & \textbf{0.5919} & \textbf{1.8038} & \textbf{27.8054} & \textbf{83.3453} \\
    \multicolumn{9}{l}{\cellcolor{gray!15}\textit{Warm records}} \\
    WAMTL~\cite{WAMTL_2024_ICWS} & 0.4897 & 1.4705 & 16.7328 & 48.2424 & 0.5242 & 1.5036 & 18.1958 & 49.4019 \\
    JQSP~\cite{JQSP_2023_TNSM}   & 0.4697 & 1.4805 & 20.7328 & 58.2424 & 0.4812 & 1.4962 & 22.1958 & 59.4019 \\
    PLHEN~\cite{PLHEN_ESA_2026}  & 0.3978 & 1.2823 & 12.4132 & \underline{41.2482} & 0.4163 & 1.2965 & 13.4075 & \underline{43.5878} \\
    MGEN~\cite{MGEN_2023_JP}     & \underline{0.3337} & 1.2518 & 12.9190 & 44.3915 & 0.3417 & 1.2298 & 13.9423 & 48.5178 \\
    DNM~\cite{DNM_2021_TSC}      & 0.3406 & \underline{1.2469} & \underline{11.8768} & 42.5943 & \underline{0.3403} & \underline{1.2285} & \underline{12.1619} & 43.6937 \\
    SHARP-QoS                    & \textbf{0.3287} & \textbf{1.2193} & \textbf{11.5223} & \textbf{41.0964} & \textbf{0.3354} & \textbf{1.2267} & \textbf{11.8719} & \textbf{43.5831} \\ \hline
    \end{tabular*}

    \smallskip
    \parbox{\textwidth}{\scriptsize
    Best results are in \textbf{bold}; the strongest baseline is \underline{underlined}.
    }
\end{table}

\subsection{Reliability of the Reported Errors}\label{sec:significance_test}
To assess the precision of the reported errors, we randomly partition the test records into $G=50$ disjoint groups of equal size and compute the MAE for each group~\cite{statistical_test}. Let $\bar{m}$ and $s$ denote the mean and sample standard deviation of these group MAEs, respectively. Using a normal approximation, the $(1-\alpha)$ confidence interval for the mean test MAE is:
\begin{equation}\label{eq:ci}
    \mathrm{CI}_{1-\alpha}=\Big[\,\bar{m}-z_{1-\alpha/2}\,\tfrac{s}{\sqrt{G}},\;\; \bar{m}+z_{1-\alpha/2}\,\tfrac{s}{\sqrt{G}}\,\Big],
\end{equation}
where $z_{1-\alpha/2}$ is the standard normal quantile, taking values 1.645, 1.960, and 2.576 for confidence levels of $90\%$, $95\%$, and $99\%$, respectively.

Table~\ref{tab:confidence_intervals} reports the mean MAE, standard deviation of the group MAEs, and corresponding confidence intervals for SHARP-QoS on WSDREAM-2T. The intervals remain relatively narrow compared with the corresponding mean MAEs, even at the $99\%$ confidence level, indicating limited uncertainty in the estimated mean errors under the normal-approximation assumption. Both the standard deviations and confidence intervals are smaller at $20\%$ training density than at $10\%$ for RT and TP, indicating lower variability among the group-level MAEs at higher density. Overall, these results support the precision of the reported mean errors under the adopted estimation procedure.

\begin{table}[!t]
    \centering
    \caption{Mean test MAE, standard deviation (STD) of group MAEs, and confidence intervals for SHARP-QoS on WSDREAM-2T ($G=50$ groups).}
    \label{tab:confidence_intervals}
    \scriptsize
    \setlength{\tabcolsep}{01pt}
    \begin{tabular*}{\columnwidth}{@{\extracolsep{\fill}}lccccc@{}}
        \hline
        Setting & MAE & STD & $90\%$ CI & $95\%$ CI & $99\%$ CI \\ \hline
        RT, TD = $10\%$ & 0.3234 & 0.141 & [0.2907,	0.3561] & [0.2844, 0.3624] & [0.2721, 0.3747] \\
        RT, TD = $20\%$ & 0.2894 & 0.122 & [0.2611, 0.3177] & [0.2557, 0.3231] & [0.2451, 0.3337] \\
        TP, TD = $10\%$ & 11.0828 & 1.976 & [10.6231, 11.5425] & [10.5351, 11.6305] & [10.3630, 11.8026] \\
        TP, TD = $20\%$ & 9.8934 & 1.722 & [9.4930, 10.2938] & [9.4163, 10.3705] & [9.2664, 10.5204] \\ \hline
    \end{tabular*}
\end{table}

\subsection{Limitations}\label{sec:limitations}
{Despite lower prediction error than the joint QoS baselines and low inference latency, SHARP-QoS has two limitations. First, evaluation covers at most 339 users and 5{,}825 services; industrial-scale scalability remains unverified despite complexity analysis (Appendix~A). Second, Cross-SNR uses a router for each ordered pair of QoS parameters, causing quadratic growth in $\mathrm{P}$ (Appendix~A). This overhead is negligible for $\mathrm{P}\le 4$ but may become substantial with more attributes. Sharing one router per source across targets could reduce router growth to linear in $\mathrm{P}$, a direction for future work.
}

\section{Related Work}\label{sec:related}
We review single-QoS and joint QoS prediction methods, followed by the multi-task learning and hyperbolic graph learning techniques underlying SHARP-QoS.

\subsection{Single-QoS Prediction}
Single-QoS methods predict one QoS parameter at a time and fall into three groups: memory-based, model-based, and hybrid. \emph{Memory-based} methods predict from user or service similarity: UPCC~\cite{upcc_1998_uai} and IPCC~\cite{ipcc_www_2001} use user- and service-based Pearson similarity, and WSRec~\cite{wsrec_2011_tsc} combines both. They are simple but degrade under sparsity and cannot represent new entities. \emph{Model-based} methods learn latent representations instead. NMF~\cite{nmf_nature_1999} and PMF~\cite{pmf_nips_2007} factorize the QoS matrix, CSMF~\cite{csmf} and GeoMF~\cite{geomf} add contextual information, CMF~\cite{cmf_www21} uses a Cauchy loss for robustness to outliers, and EFMPred~\cite{EFMPred} captures second-order interactions. Deep models such as NDMF~\cite{ndmf}, MM-DNN~\cite{MM-DNN}, and DCLG~\cite{dclg_2022_tsc} learn nonlinear interactions, while QoSGNN~\cite{QoSGNN_TSC_2024} exploits graph structure and llmQoS~\cite{llmQoS_2025_sse} uses language-model features. \emph{Hybrid} methods combine these ideas: OffDQ~\cite{offdq} joins similarity with deep networks, NCRL~\cite{ncrl_2023_tsc} fuses contextual and historical features, and ARRQP~\cite{arrqp_tsc_2025} combines QoS correlations with multi-head GCNs for anomaly-resilient prediction. All of these methods model one QoS parameter at a time, so predicting several parameters requires separate models that cannot share what they learn about the same users and services.

\subsection{Joint QoS Prediction}
Joint QoS prediction methods predict several QoS parameters with one model. DNM~\cite{DNM_2021_TSC} uses additive and multiplicative interactions between contexts, JQSP~\cite{JQSP_2023_TNSM} applies attention-based GCNs to ID-based features, and MGEN~\cite{MGEN_2023_JP} uses gated experts over location features. HTG~\cite{HTT_2024_ETT} targets cold-start prediction with graph attention but depends on AS-level topology. To limit negative transfer, PMT~\cite{PMT_2023_TNSM} and WAMTL~\cite{WAMTL_2024_ICWS} combine expert architectures with uncertainty-based and dynamic-average loss weighting, respectively, and {PLHEN~\cite{PLHEN_ESA_2026} stacks progressive hybrid expert layers with an adaptive loss.} These methods leave three gaps that SHARP-QoS addresses: they model QoS and contextual data in Euclidean space and do not exploit their hierarchical structure; they share information through increasingly large expert architectures and do not share the features learned for one QoS parameter with the others; and their loss weighting reacts to short-term loss changes, which makes task weights unstable when QoS parameters differ widely in scale. Several of them also rely on entity IDs, precise locations, or topology data, which limits privacy and cold-start applicability.

\subsection{Multi-task Learning and Hyperbolic Graph Learning}
{General MTL architectures share representations through learned combinations (Cross-Stitch~\cite{misra2016_cross-stich}), gated experts (MMoE~\cite{ma2018_MMoE}, PLE~\cite{tang2020_PLE}, DSelect-k~\cite{dselectk_NEURIPS2021}, AdaTT~\cite{adatt_kdd_2023}), or sparse sub-network routing (SNR~\cite{SNR}), which selects shared blocks for individual tasks but operates on a common input representation. Task balancing relies on loss weighting (GradNorm~\cite{gradnorm}, uncertainty weighting~\cite{Uncertainty_CVPR_2018}, DWA~\cite{dwa_2019_CVPR}, FAMO~\cite{famo_NEURIPS2023}) or gradient manipulation (MGDA~\cite{mgda}, PCGrad~\cite{pcgrad}, CAGrad~\cite{cagrad}, NashMTL~\cite{nashmtl}), which can introduce additional gradient-computation overhead. Meanwhile, hyperbolic neural networks~\cite{hyperbolic_nn_nips_2018,hyperbolic_gcn_nips_2019} can represent hierarchical graph structures with lower distortion than Euclidean embeddings, while hypergraph convolution~\cite{HGNN_AAAI2019} captures higher-order relationships. SHARP-QoS integrates these directions for QoS prediction by applying hyperbolic graph and hypergraph convolution to QoS and contextual graphs constructed exclusively from coarse public attributes, extending SNR with Cross-SNR to enable cross-parameter feature sharing, and balancing task losses through lightweight EMA-based weighting. Section~\ref{sec:experiment_section} compares SHARP-QoS with representative methods from these methodological families.
}


\section{Conclusion}\label{sec:conclusion}
This paper presents SHARP-QoS, a framework for joint QoS prediction. It learns hierarchical representations through hyperbolic graph and hypergraph convolution, facilitates feature sharing across QoS parameters and contextual sources through SNR and the proposed Cross-SNR, and balances task losses using EMA-based weighting. Together, these components enable selective cross-parameter knowledge transfer while limiting potential negative transfer. Experiments on three datasets show that SHARP-QoS achieves lower prediction errors than the evaluated joint QoS baselines across all training densities, with low inference latency. Future work will investigate scalability to large-scale service ecosystems and deployment on resource-constrained devices, while incorporating online measurements for dynamic graph updates.

\bibliographystyle{IEEEtran}
\bibliography{ref}

%


\section*{Supplementary Appendix}
Appendices A-G are provided in \url{https://github.com/csksuraj17/SHARP-QoS}

\end{document}